\documentclass{article}
\usepackage[main,final]{neurips_2026}

\usepackage[utf8]{inputenc}
\usepackage[T1]{fontenc}
\usepackage{hyperref}
\usepackage{url}
\usepackage{amsmath,amssymb,amsthm,mathtools}
\usepackage{booktabs}
\usepackage{graphicx}
\usepackage{capt-of}
\usepackage{xcolor}
\usepackage{microtype}
\usepackage{algorithm}
\usepackage{algpseudocode}
\usepackage[shortlabels]{enumitem}

\hypersetup{colorlinks=true,linkcolor=blue!55!black,citecolor=blue!55!black,urlcolor=blue!55!black}

\theoremstyle{plain}
\newtheorem{theorem}{Theorem}[section]
\newtheorem{proposition}[theorem]{Proposition}
\newtheorem{lemma}[theorem]{Lemma}
\newtheorem{corollary}[theorem]{Corollary}
\theoremstyle{definition}
\newtheorem{definition}[theorem]{Definition}
\newtheorem{assumption}[theorem]{Assumption}
\newtheorem{example}[theorem]{Example}
\theoremstyle{remark}
\newtheorem{remark}[theorem]{Remark}

\newcommand{\R}{\mathbb{R}}
\newcommand{\E}{\mathbb{E}}
\newcommand{\N}{\mathcal{N}}
\newcommand{\tr}{\operatorname{tr}}
\newcommand{\diag}{\operatorname{diag}}
\newcommand{\cov}{\operatorname{cov}}

\newcommand{\Rk}{\mathcal{R}}
\newcommand{\dnu}{\,d\nu(\rho)}
\newcommand{\Shat}{\hat S}
\newcommand{\iso}{\mathrm{iso}}
\newcommand{\pin}{\mathcal{P}}
\newcommand{\rbldc}{RB-LDC}

\newcommand{\includefig}[2]{\IfFileExists{#2}{\includegraphics[#1]{#2}}{%
  \fbox{\parbox[c][0.18\textheight][c]{0.9\linewidth}{\centering\small missing file: \texttt{\detokenize{#2}}}}}}

\title{RB-LDC: Redundancy-Balanced Latent Coding\\for Noise-Robust Diffusion Latents}

\author{%
  Hyunseok Jeong \\
  DGIST \\
  Daegu, Republic of Korea \\
  \texttt{hsj@dgist.ac.kr}
  \And
  Jaeho Jeon \\
  DGIST \\
  Daegu, Republic of Korea \\
  \texttt{dgwogh@dgist.ac.kr}
  \And
  Young-Sik Kim\thanks{Corresponding author.} \\
  DGIST \\
  Daegu, Republic of Korea \\
  \texttt{ysk@dgist.ac.kr}
}

\begin{document}
\maketitle

\begin{abstract}
Latent diffusion models decode generated latents that are only approximately correct, so the sensitivity of the decoder to latent perturbations affects the images they produce. After whitening, the forward diffusion process corrupts every latent direction at the same signal-to-noise ratio (SNR), although a decoder can be far more sensitive to some directions than to others. We study an invertible linear layer inserted between the encoder and the diffusion process. In the eigenbasis of its Gram matrix, this layer shifts the log-SNR of each latent direction, and its effect on linear minimum mean-square-error (LMMSE) recovery admits an exact spectral expression for any whitened latent distribution. Minimizing the decoder-sensitivity-weighted LMMSE risk, integrated over the diffusion schedule under a latent-energy budget with a floor that keeps the layer invertible, gives a unique solution that a deterministic spectral solver computes: the optimal layer aligns with the decoder's expected pullback metric and equalizes the marginal reduction of risk across directions. In the large-budget limit this solution becomes a power law in sensitivity whose exponent is set by the low-SNR tail of the loss weighting, one half for noise prediction and two thirds for data prediction under the cosine schedule, with a crossover between the two on a truncated schedule. We instantiate this principle as \rbldc{}, which requires only a Jacobian-based estimate of decoder sensitivity. On ImageNet $256\times256$ with the SD-VAE and VA-VAE tokenizers, our trained implementation of \rbldc{} improves reconstruction from noisy latents over budget-matched isotropic coding in all four metrics and all five coding configurations when averaged over 15 SNRs and three pipeline seeds, reducing the average perturbed FID by 17.1 on VA-VAE block-local coding, while leaving clean reconstruction unchanged; we also report the intermediate SNRs at which it degrades perturbed FID. With a frozen diffusion model, \rbldc{} also yields lower FID point estimates in all 12 tested combinations of sampling progress and channel SNR during sampling.
\end{abstract}

\section{Introduction}
\label{sec:intro}

Latent diffusion models (LDMs) train a diffusion model in the latent space of a pretrained autoencoder, or tokenizer, and map generated latents back to images with its decoder \citep{rombach2022ldm}. Generated latents are never exact, so the decoder operates on perturbed inputs. Recent work finds that the robustness of the latent space to such perturbations is closely related to generation quality and proposes tokenizer training objectives that improve it \citep{qiu2025robusttok,li2026veloss}; related design choices, such as training the decoder on noise-augmented latents and shifting the noise schedule with the latent dimension, are also important in recent systems \citep{zheng2026rae}.

This paper studies a degree of freedom that lies between the tokenizer and the diffusion process. After whitening, the forward process $Z_t=\alpha_tZ_0+\sigma_t\varepsilon$ corrupts every latent direction at the same signal-to-noise ratio (SNR) $\rho_t=\alpha_t^2/\sigma_t^2$, whereas the decoder may be much more sensitive to some directions than to others; for the VA-VAE tokenizer \citep{yao2025vavae}, the sensitivity matrix of a latent block has condition number 408 in our measurements. In an oracle experiment with a frozen diffusion transformer (DiT), correcting its one-step clean-latent predictions along high-sensitivity directions yields a larger average reduction in decoded-image mean-square error than correcting low-sensitivity directions with the same reduction in whitened latent error energy (Table~\ref{tab:dit-error-correction}). This observation motivates accounting for decoder sensitivity when allocating protection across latent directions. An invertible linear layer $G$ applied before the forward process allows direction-dependent SNRs. In the eigenbasis of $G^\top G$ with eigenvalues $\lambda_i$, direction $i$ is observed at SNR $\rho_t\lambda_i$, so $G$ shifts the log-SNR of each direction by $\log\lambda_i$. Scaling the input by a constant, which shifts the log-SNR of all directions by the same amount, is an effective way to adapt the noise schedule \citep{chen2023noise,hoogeboom2023simple,esser2024sd3}, and forward processes with component-wise noise levels have been studied in Fourier and other bases \citep{gerdes2024gud,falck2025fourier}. We ask how the SNR should be distributed across latent directions when the total latent energy is fixed.

We answer this question with an estimation criterion: the linear minimum mean-square error (LMMSE) of the whitened latent, weighted by the expected pullback metric of the decoder and integrated over the diffusion schedule. For a single SNR, this criterion reduces to the weighted mean-square-error design of linear precoders in communications \citep{leepetersen1976,scaglione2002,palomar2003}. Integrating over the schedule changes the solution in a way that has no single-SNR counterpart: the low-SNR tail of the loss-weighted SNR measure determines how steeply the optimal allocation grows with sensitivity. We call the resulting design \rbldc{} (redundancy-balanced latent coding), where \emph{redundancy} refers to the protection energy $\lambda_i$ assigned to each latent direction. Our contributions are as follows.
\begin{itemize}[leftmargin=*,itemsep=1pt,topsep=2pt]
\item \textbf{Main theoretical result.} The exponent of the optimal allocation is not a property of decoder sensitivity alone: it is set by the low-SNR tail of the weighting that the diffusion objective places on noise levels (Theorem~\ref{thm:exponent}). On the cosine schedule the exponent is $1/2$ for noise prediction and $2/3$ for data prediction, and a schedule truncated at $\rho_{\min}$ produces a crossover between the two (Corollary~\ref{cor:cosine}).
\item \textbf{Supporting structure.} The optimal layer is aligned with the decoder pullback metric (Theorem~\ref{thm:align}) and equalizes marginal risk reduction across directions, with a unique solution under a floor that keeps the layer invertible (Theorem~\ref{thm:balance}); the attainable gain over isotropic coding is bounded by the condition number of the sensitivity matrix (Proposition~\ref{prop:gain}). We also show that the risk depends on the code only through the Gram matrix $G^\top G$ (Proposition~\ref{prop:spectrum}), and only through its spectrum once the alignment of Theorem~\ref{thm:align} is imposed, and that an imperfectly whitened latent reduces to the same problem exactly (Remark~\ref{rem:colored}).
\item \textbf{Method.} \rbldc{} estimates decoder sensitivity with vector--Jacobian products and computes the allocation with a deterministic spectral solver. Its risk degrades by at most the sandwich ratio of the estimate (Proposition~\ref{prop:sandwich}), and block-diagonal codes admit an exact risk decomposition (Lemma~\ref{lem:block}).
\item \textbf{Experiments.} On ImageNet with SD-VAE and VA-VAE, \rbldc{} improves the SNR-averaged perturbed FID, PSNR, SSIM, and LPIPS of reconstructions from noisy latents in all five coding configurations across three pipeline seeds, with the largest gains for block-local coding on VA-VAE, whose sensitivity matrix is more anisotropic than that of SD-VAE. We also evaluate corruption during sampling and latent transmission after generation with a frozen diffusion model.
\end{itemize}
The optimized risk is a surrogate, and we delimit it: Proposition~\ref{prop:excess} makes the gap to a trained estimator explicit, and the surrogate's link to nonlinear feature distortion is asymptotically tight under noise-prediction weighting but not under uniform weighting (Appendix~\ref{app:nonlinear}). A broader discussion of related work is provided in Appendix~\ref{sec:related}.

\section{Latent coding as SNR allocation}
\label{sec:setup}

\paragraph{Coded latents and the diffusion channel.}
Let $S\in\R^k$ be the latent of an image after a fixed whitening transform, so that $\E S=0$ and $\E SS^\top=I_k$; no further distributional assumption is made unless stated. A coding layer $G\in\R^{m\times k}$ with $m\ge k$ produces $Z_0=GS$, which is corrupted by the forward process \citep{ho2020ddpm,kingma2021vdm}
\begin{equation}
Z_t=\alpha_tZ_0+\sigma_t\varepsilon,\qquad \varepsilon\sim\N(0,I_m),\qquad \rho_t=\alpha_t^2/\sigma_t^2 .
\label{eq:channel}
\end{equation}
We write $M=G^\top G$ and fix the latent energy $\E\|Z_0\|^2=\tr M\le B$. The isotropic code $M=(B/k)I_k$ spreads the budget uniformly. A diffusion objective weights noise levels by a function $w(t)$ \citep{kingma2023elbo}; we summarize this weighting by the \emph{SNR measure} $\nu$, the push-forward of $w(t)\,dt$ under $t\mapsto\rho_t$, which is a finite sum of point masses for a discrete schedule.

\paragraph{Linear recovery.}
Recovery is analyzed with the linear minimum mean-square-error estimator, which needs no assumption beyond second moments.

\begin{proposition}[LMMSE recovery]
\label{prop:lmmse}
Let $Y=Z_t/\sigma_t=\sqrt{\rho}\,GS+\varepsilon$ with $\rho=\rho_t$.
(a) For any $S$ with $\E S=0$ and $\E SS^\top=I_k$, the best linear estimator of $S$ from $Y$ is $\Shat=\sqrt{\rho}\,(I_k+\rho M)^{-1}G^\top Y$, and its error covariance is $\Sigma_\rho(M)=(I_k+\rho M)^{-1}$.
(b) If $S$ is Gaussian, then $\Shat=\E[S\mid Y]$ and $\Sigma_\rho(M)$ is the posterior covariance; for general $S$, the error covariance of $\E[S\mid Y]$ is dominated by $\Sigma_\rho(M)$ in the positive semidefinite order.
\end{proposition}

\paragraph{Sensitivity-weighted risk.}
Let $D$ denote the decoder composed with the inverse whitening and let $\Phi$ be a fixed feature map, such as the identity or a pretrained network. With $F=\Phi\circ D$ and Jacobian $J_F$, the matrix $H(s)=J_F(s)^\top J_F(s)$ is the pullback of the Euclidean feature metric to the latent space \citep{arvanitidis2018latent}, and $H=\E H(S)$ is its average. We write $H=U\diag(h_1,\dots,h_k)U^\top$ with $h_1\ge\dots\ge h_k>0$.

\begin{definition}[Integrated risk]
\label{def:risk}
For $M\succeq0$, let $\Rk(M)=\int\tr\big(H\,\Sigma_\rho(M)\big)\dnu=\tr\big(H\varphi(M)\big)$, where $\varphi(\lambda)=\int(1+\rho\lambda)^{-1}\dnu$ is applied to $M$ through its spectrum.
\end{definition}

By Proposition~\ref{prop:lmmse}, $\Rk(M)$ is exactly the $H$-weighted LMMSE error integrated over the schedule. For a Gaussian latent and a decoder with a Lipschitz Jacobian, it is also the leading term of the expected feature distortion $\E\|F(S)-F(\Shat)\|^2$, with a remainder of order $(\tr\Sigma_\rho)^{3/2}$ (Proposition~\ref{prop:feature} in Appendix~\ref{app:nonlinear}).

\begin{remark}[Imperfect whitening]
\label{rem:colored}
If the latent has covariance $C\succ0$ rather than $I_k$, all statements apply after the change of variables $\tilde S=C^{-1/2}S$, with $\tilde G=GC^{1/2}$, $\tilde M=\tilde G^\top\tilde G$, and $\tilde H=C^{1/2}HC^{1/2}$. The constraints move with it: the energy is $\tr(MC)=\tr\tilde M$, the floor $M\succeq\ell I_k$ becomes $\tilde M\succeq\ell C$, and block-diagonal codes stay block-diagonal only if $C$ is block-diagonal for the same partition, so we state budget, floor, and partition in the whitened coordinates. If $C$ is not corrected, the estimator of Proposition~\ref{prop:lmmse} has error covariance $A_\rho(C+\rho M)A_\rho$ with $A_\rho=(I_k+\rho M)^{-1}$, and for $\|C-I_k\|_{\mathrm{op}}\le\tau<1$ its risk is within a factor $1\pm\tau$ of $\Rk(M)$, so the white-model design is within $(1+\tau)/(1-\tau)$ of optimal (Proposition~\ref{prop:colored}).
\end{remark}

\paragraph{The design variable.}
\begin{proposition}[Gram-matrix dependence and dimension invariance]
\label{prop:spectrum}
Under \eqref{eq:channel}, $\Sigma_\rho$ and $\Rk$ depend on $G$ only through $M=G^\top G$, and every $M\succeq0$ with $\tr M\le B$ is attained by a square code ($m=k$). Hence the minimal risk is the same for all $m\ge k$, and left multiplication of $G$ by an orthogonal matrix does not change $\Rk$. The risk is not determined by the spectrum of $M$ alone: it also depends on the orientation of the eigenvectors of $M$ relative to $H$, which Theorem~\ref{thm:align} resolves.
\end{proposition}

At a fixed total trace budget under isotropic Gaussian channel noise, increasing $m$ beyond $k$ cannot reduce the modeled LMMSE risk. Per-coordinate constraints, erasures, non-isotropic noise, or restrictions on a trained receiver define different design problems. Within our model, the design distributes energy $B$ across latent directions, with the SNR measure determined by the diffusion objective.

\begin{lemma}[Diffusion losses on coded latents]
\label{lem:loss}
Let $G$ be invertible and $\alpha_t^2+\sigma_t^2=1$, and let the output of a denoiser be parameterized by an estimate $\hat s$ of $s$, with error $\delta=\hat s-s$. The noise-, data-, and velocity-prediction losses, with $v=\alpha_t\varepsilon-\sigma_tz_0$ \citep{salimans2022progressive}, satisfy
\begin{equation*}
\|\varepsilon-\hat\varepsilon\|^2=\rho_t\,\delta^\top M\delta,\qquad
\|z_0-\hat z_0\|^2=\delta^\top M\delta,\qquad
\|v-\hat v\|^2=(1+\rho_t)\,\delta^\top M\delta.
\end{equation*}
With uniform time sampling, these objectives therefore correspond to $w(t)=\rho_t$, $1$, and $1+\rho_t$, respectively \citep{ho2020ddpm,hang2023minsnr}.
\end{lemma}

We take $w$ in Definition~\ref{def:risk} to be the weighting of the diffusion objective in use. Only the weighting is taken from that objective: by Lemma~\ref{lem:loss} the diffusion loss weights direction $i$ by $\lambda_i$, whereas $\Rk$ weights it by decoder sensitivity and fixes the estimator to be linear. Thus $\Rk$ is a decoder-aware reconstruction surrogate that borrows the time weighting of the objective, not the objective itself.

The risk $\Rk$ is the error of the LMMSE estimator. For a trained estimator the two contributions separate exactly.

\begin{proposition}[Excess risk of a trained estimator]
\label{prop:excess}
Let $\Shat_{\mathrm B}=\E[S\mid Y]$ with error covariance $\Sigma^{\mathrm B}_\rho(M)$, let $\Rk_{\mathrm B}(M)=\int\tr(H\Sigma^{\mathrm B}_\rho(M))\dnu$, and let $\mathcal A(M)=\Rk(M)-\Rk_{\mathrm B}(M)\ge0$, which vanishes for Gaussian $S$. For any square-integrable estimator $\Shat_\theta=f(Y)$, with $\mathcal E_\theta(M)=\int\E\|H^{1/2}(\Shat_{\mathrm B}-\Shat_\theta)\|^2\dnu$,
\begin{equation}
\mathcal L_\theta(M):=\int\E\big\|H^{1/2}(S-\Shat_\theta)\big\|^2\dnu=\Rk(M)-\mathcal A(M)+\mathcal E_\theta(M) .
\label{eq:excess}
\end{equation}
Hence $f$ prefers $M$ to $M'$ exactly when $\mathcal E_\theta(M)-\mathcal E_\theta(M')-\big(\mathcal A(M)-\mathcal A(M')\big)<\Rk(M')-\Rk(M)$. For Gaussian $S$ the term $\mathcal A$ vanishes; for a general whitened latent it does not, and omitting it can reverse the comparison (Example~\ref{ex:bayesgap}).
\end{proposition}

\begin{remark}[Scope]
\label{rem:scope}
In the idealized limit of exact scores and exact integration of the reverse process, an invertible $G$ does not change the distribution of generated latents, in line with the invariance of continuous-time diffusion models to the specification of the forward process \citep{kingma2021vdm}; convergence guarantees for diffusion sampling are expressed through score-estimation and discretization errors \citep{chen2023sampling}. The effect of $G$ on a trained model is therefore governed by $\mathcal E_\theta$ and $\mathcal A$ in~\eqref{eq:excess}, neither of which we model. Our analysis isolates the channel term through LMMSE recovery from noisy latents. Section~\ref{sec:exp} additionally evaluates this recovery rule for added corruption during sampling with a frozen diffusion model and for latent transmission after generation (Appendix~\ref{app:results}). Training a DiT on the coded representations increases clean-generation FID relative to vanilla at 150k and 400k updates in the runs of Table~\ref{tab:clean-generation}.
\end{remark}

\section{Optimal allocation}
\label{sec:theory}

We assume that $\nu$ is a nonzero finite measure supported in $[\rho_{\min},\rho_{\max}]\subset(0,\infty)$ unless stated otherwise, and we define $m(\lambda)=-\varphi'(\lambda)=\int\rho\,(1+\rho\lambda)^{-2}\dnu$, the marginal reduction of risk per unit of energy at level $\lambda$. Both $\varphi$ and $m$ are strictly decreasing, and $\varphi$ is strictly convex (Lemma~\ref{lem:kernels}).

\begin{theorem}[Alignment]
\label{thm:align}
For every $M\succeq0$ with eigenvalues $\lambda_1\ge\dots\ge\lambda_k$, $\Rk(M)\ge\sum_ih_i\varphi(\lambda_i)$, with equality if $M=U\diag(\lambda)U^\top$. Consequently, for any floor $\ell\ge0$ with $B>k\ell$, minimizing $\Rk$ subject to $\tr M\le B$ and $M\succeq\ell I_k$ is equivalent to
\begin{equation}
\min_{\lambda\in\R^k}\ \sum_{i=1}^kh_i\,\varphi(\lambda_i)\qquad\text{subject to}\qquad\lambda_i\ge\ell,\quad\sum_{i=1}^k\lambda_i\le B .
\label{eq:scalar}
\end{equation}
\end{theorem}

\begin{theorem}[Redundancy balance]
\label{thm:balance}
Problem~\eqref{eq:scalar} has a unique solution $\lambda^\star$, and $\sum_i\lambda_i^\star=B$. There is $\mu>0$ such that
\begin{equation}
h_i\,m(\lambda_i^\star)=\mu\ \ \text{if}\ \lambda_i^\star>\ell,\qquad h_i\,m(\ell)\le\mu\ \ \text{if}\ \lambda_i^\star=\ell .
\label{eq:kkt}
\end{equation}
Moreover, $h_i\ge h_j$ implies $\lambda_i^\star\ge\lambda_j^\star$, and $\lambda_i^\star>\ell$ for every $i$ when $B$ is sufficiently large.
\end{theorem}

At the optimum, each direction above the floor receives energy until its marginal risk reduction $h_i\,m(\lambda_i)$ equals that of every other such direction. Because $m$ depends on the entire SNR measure, the shape of the allocation depends on the schedule and on the loss weighting.

For $\ell>0$, the floor makes the layer invertible: $\lambda_{\min}(G^\top G)\ge\ell$ gives $\mathrm{cond}(G)\le\sqrt{\lambda_1^\star/\ell}$ and exact inversion on clean latents. With $\ell=0$ the solution can be singular; for $H=\diag(100,1)$, $\nu=\delta_1$ and $B=2$ it is $\lambda^\star=(2,0)$, so the layer is not invertible and the deployment of Section~\ref{sec:method} does not apply (Example~\ref{ex:singular}). All results below hold for every $\ell\ge0$, and the floor is inactive once $B$ is large enough, so it does not affect the asymptotic statements.

\begin{theorem}[Allocation exponent]
\label{thm:exponent}
Let $\beta\in(1,2]$ be defined by one of the following conditions.
(a) If $\int\rho^{-1}\dnu<\infty$, let $\beta=2$.
(b) If $\nu$ is a finite measure on $(0,\rho_{\max}]$ with $\nu(d\rho)=C\rho^{\gamma-1}(1+o(1))\,d\rho$ as $\rho\to0$ for some $C,\gamma>0$, let $\beta=\min(2,1+\gamma)$.
Then, as $B\to\infty$,
\begin{equation}
\lambda_i^\star(B)=B\,\frac{h_i^{1/\beta}}{\sum_jh_j^{1/\beta}}\,\big(1+o(1)\big),
\label{eq:exponent}
\end{equation}
with a logarithmically slow rate when $\gamma=1$. In particular, case (a) gives the square-root law $\lambda_i^\star\propto\sqrt{h_i}$.
\end{theorem}

\begin{corollary}[Cosine schedule and loss weighting]
\label{cor:cosine}
Let $\rho_t=\cot^2(\pi t/2)$, the cosine schedule of \citet{nichol2021improved} without its offset, with $t$ uniform on $(t_0,1)$ for some $t_0>0$, and let $w(t)=\rho_t^{\,a}(1+o(1))$ as $\rho_t\to0$ for some $a\ge0$. Then $\gamma=a+\frac12$ and $\beta=\min(2,a+\frac32)$. Consequently, by Lemma~\ref{lem:loss}, the noise-prediction objective and the min-SNR weighting of \citet{hang2023minsnr} ($a=1$) lead to $\lambda_i^\star\propto h_i^{1/2}$, whereas the data- and velocity-prediction objectives ($a=0$) lead to $\lambda_i^\star\propto h_i^{2/3}$. For a schedule truncated at $\rho_{\min}>0$ with $a<1/2$, the allocation ratios are close to $(h_i/h_j)^{1/\beta}$ when all $\lambda_i^\star$ lie in $[\Lambda,\epsilon\rho_{\min}^{-1}]$, with $\Lambda$ sufficiently large and $\epsilon$ sufficiently small (Corollary~\ref{cor:truncated}). For every $a\ge0$, the large-budget limit on a truncated schedule is the square-root law, in the regime $\min_i\lambda_i^\star\gg\rho_{\min}^{-1}$.
\end{corollary}

\begin{proposition}[Gain over isotropic coding]
\label{prop:gain}
Let $\lambda^\iso=(B/k)\mathbf{1}$, $\kappa=h_1/h_k$, and $\Gamma(h)=k\sum_ih_i/(\sum_i\sqrt{h_i})^2$.
(a) If $\int\rho^{-1}\dnu<\infty$, then $\Rk(\lambda^\iso)/\Rk(\lambda^\star)\to\Gamma(h)$ as $B\to\infty$.
(b) $1\le\Gamma(h)\le(1+\sqrt\kappa)^2/(4\sqrt\kappa)$, and $\Gamma(h)=1$ if and only if all $h_i$ are equal.
(c) The per-SNR risk difference $g(\rho)=\sum_ih_i\big[(1+\rho B/k)^{-1}-(1+\rho\lambda_i^\star)^{-1}\big]$ is nonnegative for all sufficiently small $\rho$ and, when $B$ is large enough, for all sufficiently large $\rho$.
\end{proposition}

The bound in (b) increases with the condition number of $H$; it equals $1.67$ for $\kappa=20$ and $5.56$ for $\kappa=408$. It is only a bound: with $k=128$, the spectrum $(408,\dots,408,1)$ gives $\Gamma=1.01$ while $(20,\dots,20,1,\dots,1)$ with $64$ entries of each value gives $\Gamma=1.40$, so a larger condition number does not imply a larger gain, and $\Gamma$ must be computed from the whole spectrum.

Theorem~\ref{thm:exponent} is asymptotic, whereas deployments use a finite budget. The following exact statement describes the allocation at any budget and recovers the exponent in the limit.

\begin{proposition}[Local exponent]
\label{prop:local}
Let $\beta_{\mathrm{eff}}(\lambda)=-\,d\log m(\lambda)/d\log\lambda=2\lambda\int\rho^2(1+\rho\lambda)^{-3}\dnu\,\big/\int\rho(1+\rho\lambda)^{-2}\dnu\in(0,2]$. Along the balance curve $h\,m(\lambda)=\mu$ at fixed $\mu$, which~\eqref{eq:kkt} imposes across directions, $d\log\lambda/d\log h=1/\beta_{\mathrm{eff}}(\lambda)$; this compares directions at a common multiplier and is not the effect of changing one $h_i$ at a fixed budget, which also moves $\mu$. Consequently, if $h_i\ge h_j$, both coordinates exceed the floor, and $\beta_-\le\beta_{\mathrm{eff}}\le\beta_+$ on $[\lambda_j^\star,\lambda_i^\star]$, then
\begin{equation}
\big(h_i/h_j\big)^{1/\beta_+}\ \le\ \lambda_i^\star/\lambda_j^\star\ \le\ \big(h_i/h_j\big)^{1/\beta_-} .
\label{eq:localexp}
\end{equation}
Moreover $\beta_{\mathrm{eff}}(\lambda)\to\beta$ as $\lambda\to\infty$, with $\beta$ as in Theorem~\ref{thm:exponent}.
\end{proposition}

\begin{remark}[Relation to linear precoding and to the diffusion loss]
\label{rem:precoding}
For a point mass $\nu=\delta_\rho$, problem~\eqref{eq:scalar} with $\ell=0$ is the weighted mean-square-error power allocation of linear transceivers, with solution $\lambda_i^\star=[\sqrt{h_i/(\mu\rho)}-1/\rho]_+$ \citep{leepetersen1976,scaglione2002,palomar2003,xing2013weighted}; Theorem~\ref{thm:align} is the corresponding alignment result. Averaging the mean-square error over a channel distribution rather than a single SNR leads to the same marginal-balancing structure, with the allocation obtained by inverting the averaged kernel, as in linear transmission over a fading channel \citep{tan2016linear}. What is specific to the diffusion setting is that this averaging measure is the loss weighting of the training objective: Theorem~\ref{thm:exponent} and Corollary~\ref{cor:cosine} show that its low-SNR tail fixes the exponent, and Proposition~\ref{prop:local} describes the allocation at a finite budget. By Lemma~\ref{lem:loss}, a diffusion model trained on $GS$ weights direction $i$ of its loss by $\lambda_i$, so at the allocation of Theorem~\ref{thm:exponent} the loss weights directions approximately in proportion to $h_i^{1/\beta}$.
\end{remark}

\section{RB-LDC}
\label{sec:method}

\paragraph{Estimating decoder sensitivity.}
For a probe $u\sim\N(0,I_p)$, the vector--Jacobian product $v=J_F(s)^\top u$, computed by backpropagation through the frozen decoder, satisfies $\E_u\,vv^\top=H(s)$. Averaging $vv^\top$ over latents and probes therefore gives an unbiased estimate $\hat H$ without forming Jacobians (Proposition~\ref{prop:vjp}). The allocation is robust to estimation error in the following sense; Appendix~\ref{app:estimation} gives a sufficient number of probes under a Gaussian model.

\begin{proposition}[Robustness to the sensitivity estimate]
\label{prop:sandwich}
Suppose that $aH\preceq\hat H\preceq bH$ for some $0<a\le b$. If $\hat M$ minimizes the risk computed with $\hat H$ and $M^\star$ minimizes the risk computed with $H$ over the same feasible set, then $\Rk_H(\hat M)\le(b/a)\,\Rk_H(M^\star)$.
\end{proposition}

\paragraph{Deterministic spectral solver.}
Algorithm~\ref{alg:rbldc} returns the solution of Theorem~\ref{thm:balance} with one eigendecomposition and a one-dimensional search, because $m$ is strictly decreasing and $\sum_i\lambda_i(\mu)$ is monotone in $\mu$; for a general $\nu$ both $m^{-1}$ and $\mu$ are computed numerically, so we call it a deterministic spectral solver and reserve the term \emph{closed form} for the asymptotic power law of Theorem~\ref{thm:exponent}. The layer can also be trained by gradient descent with $G=Q\diag(\sqrt\lambda)P^\top$, where $P,Q$ are orthogonal through the Cayley transform and $\lambda=\ell\mathbf 1+(B-k\ell)\,\mathrm{softmax}(u)$, which keeps the budget exact and suits joint fine-tuning; it covers the relative interior of the feasible set, so allocations with $\lambda_i=\ell$ are approached but not attained. The main reconstruction experiments in Section~\ref{sec:exp} use this trained form with $\ell=0$, optimized with Adam \citep{kingma2015adam}: its spectrum is strictly positive, hence the code is invertible, and on the primary configuration it matches Algorithm~\ref{alg:rbldc} to a relative error of $0.005$ (Table~\ref{tab:ablation}).

\begin{algorithm}[htbp]
\caption{\rbldc{} allocation}
\label{alg:rbldc}
\begin{algorithmic}[1]
\Require sensitivity estimate $\hat H$, SNR measure $\nu$, budget $B$, floor $\ell\ge0$ with $B>k\ell$
\State $\hat H=U\diag(\hat h)U^\top$ \Comment{eigendecomposition}
\State find $\mu>0$ by bisection such that $\sum_i\lambda_i(\mu)=B$, where $\lambda_i(\mu)=m^{-1}(\mu/\hat h_i)$ if $\hat h_i\,m(\ell)>\mu$ and $\lambda_i(\mu)=\ell$ otherwise
\State \Return $G=\diag\big(\sqrt{\lambda(\mu)}\big)\,U^\top$ \Comment{invertible for $\ell>0$}
\end{algorithmic}
\end{algorithm}

\paragraph{Block-local coding.}
Tokenizer latents are spatial maps with $C$ channels. Channel-mix coding applies a $C\times C$ code at every spatial position ($k=C$), and block-local coding applies a code of size $k=Cb^2$ to each non-overlapping $b\times b$ block. Both are block-diagonal codes on the full latent, for which the risk decomposes exactly.

\begin{lemma}[Block-diagonal codes]
\label{lem:block}
Let $M=\diag(M_1,\dots,M_L)$ in a fixed partition, and let $H_{bb}$ denote the corresponding diagonal blocks of $H$. Then $\Rk(M)=\sum_b\tr(H_{bb}\varphi(M_b))$ for every $H$; if all blocks share $M_0$, then $\Rk(M)=\tr(H_{\mathrm{eff}}\varphi(M_0))$ with $H_{\mathrm{eff}}=\sum_bH_{bb}$. Under a single budget, the best block-diagonal code is given by Theorem~\ref{thm:balance} applied jointly to the eigenvalues of all $H_{bb}$, and its excess risk over the best unrestricted code $M^\star$ is at most $-\tr(H_{\mathrm{OD}}\varphi(M^\star))$, where $H_{\mathrm{OD}}$ is the off-block part of $H$.
\end{lemma}

The off-block entries of $H$ therefore affect only the gap to the unrestricted optimum and never the risk of a given block-diagonal code. The bound compares the best block-diagonal code, whose blocks may differ, with the unrestricted optimum. Using a single shared code across all blocks is a further restriction whose cost the bound does not cover; that cost vanishes whenever a single code attains the pooled per-block optimum, for which identical $H_{bb}$ are sufficient but not necessary (Appendix~\ref{app:block}).

\paragraph{Deployment.}
\rbldc{} whitens the latent and applies $G$, and the diffusion process operates on $Z_0=GS$; a latent is mapped back by $G^{-1}$, the inverse whitening, and the decoder. An invertible $G$ preserves clean reconstruction up to floating-point error; the positive-floor solution additionally satisfies $\mathrm{cond}(G)\le\sqrt{\lambda_1^\star/\ell}$ for $\ell>0$. For reconstruction from noisy latents (Section~\ref{sec:exp}), $G^{-1}$ is replaced by the estimator of Proposition~\ref{prop:lmmse}. The added computation is one $k\times k$ matrix multiplication per block.

\section{Experiments}
\label{sec:exp}

\paragraph{Setup.}
For reconstruction from noisy latents, we use 5,000 images of the ImageNet validation set \citep{deng2009imagenet} at $256\times256$ resolution and two frozen tokenizers: the Stable Diffusion autoencoder (SD-VAE; \citealp{rombach2022ldm}) and VA-VAE \citep{yao2025vavae}. We evaluate five coding configurations: channel-mix coding on SD-VAE ($k=4$) and VA-VAE ($k=32$), block-local coding with $b=2$ on SD-VAE ($k=16$) and VA-VAE ($k=128$), and block-local coding with $b=4$ on VA-VAE ($k=512$). The primary configuration is VA-VAE block-local coding with $k=128$. Latents are whitened with a calibrated transform, whose residual $\|\cov(S)-I_k\|_F/\sqrt k$ is $0.057$ for SD-VAE channel-mix coding and $0.066$ for VA-VAE block-local coding. The baseline is the isotropic code in the same coordinates; \rbldc{} uses the same budget $B=k$ and SNR weighting $w(t)=\rho_t$ on a cosine schedule with 1,000 steps, and estimates $H$ with $\Phi$ equal to the identity unless stated otherwise, the alternatives being a frozen randomly initialized convolutional network \citep{he2015delving} and VGG-16 features \citep{simonyan2015vgg}.

For each of 15 SNRs, $\log_{10}\rho\in\{-4.0,-3.5,\dots,+3.0\}$, we add Gaussian noise to the coded latent, recover $\Shat$ with the estimator of Proposition~\ref{prop:lmmse}, and decode. Within each run, the noise realization at each SNR is fixed and shared by all allocations, so comparisons are paired. We report the Fr\'echet Inception distance \citep{heusel2017ttur,szegedy2016inception} between the 5,000 reconstructions and the original images, which we call the perturbed FID (pFID) following \citet{qiu2025robusttok}, who introduced a perturbed FID for tokenizers under a different perturbation model; we also report per-image PSNR, SSIM \citep{wang2004ssim}, and LPIPS with VGG features \citep{zhang2018lpips}, averaged over images. The FID of clean reconstructions (rFID) is an implementation check. For this reconstruction study, all differences are \rbldc{} minus isotropic coding, and averages over the 15 SNRs are unweighted averages over a uniform grid in $\log_{10}\rho$, which summarize the sweep and are not the design objective $\Rk$. Appendix~\ref{app:details} lists all details.

\paragraph{Decoder sensitivity and DiT prediction errors.}
We test whether decoder sensitivity matters for the reconstruction errors of a trained denoiser using a frozen LightningDiT-XL/1 checkpoint (400k updates, CFG 1.0) with VA-VAE \citep{peebles2023dit,yao2025vavae}. On 256 ImageNet validation images, we obtain one-step clean-latent predictions from forward-corrupted latents at five linear-flow times, with $t=0$ corresponding to noise and $t=1$ to clean data. Using the primary identity-$\Phi$ estimate $\hat H$, we partially remove the prediction error in either its top or bottom 16 eigendirections in each of the 64 whitened blocks ($k=128$). For each image and time, the two corrections share the same prediction and reduce total whitened error energy by the same amount, set to half the smaller of the two projected error energies. Appendix~\ref{app:details} specifies the correction protocol.

Table~\ref{tab:dit-error-correction} reports image MSE relative to the clean decoded image. At $t=0.5$, the mean removed fraction of latent error energy is $2.72\%$, while mean image MSE decreases by $18.96\%$ for the top directions and $0.33\%$ for the bottom directions. This oracle intervention uses the ground-truth latent and tests the relevance of decoder sensitivity to reconstruction from DiT predictions; it does not evaluate \rbldc{} performance or generation quality.

\begin{table}[htbp]
\caption{Oracle correction of DiT predictions on 256 images (seed 42). High and Low correct the top and bottom 16 eigendirections of the primary identity-$\Phi$ sensitivity estimate. Removed energy is the mean percentage of total whitened error energy removed per image, matched between corrections to numerical precision. Image MSE is averaged over images against their clean reconstructions; lower is better.}
\label{tab:dit-error-correction}
\centering
\small
\setlength{\tabcolsep}{5pt}
\begin{tabular}{@{}ccccc@{}}
\toprule
& & \multicolumn{3}{c}{Image MSE $\downarrow$}\\
\cmidrule(lr){3-5}
$t$ & Removed energy (\%) & Original & High & Low\\
\midrule
$0.10$ & $3.19$ & $0.238641$ & $0.156636$ & $0.240446$\\
$0.25$ & $3.09$ & $0.168840$ & $0.121811$ & $0.168581$\\
$0.50$ & $2.72$ & $0.099320$ & $0.080486$ & $0.098988$\\
$0.75$ & $2.18$ & $0.042703$ & $0.038128$ & $0.042549$\\
$0.90$ & $1.86$ & $0.012755$ & $0.011873$ & $0.012677$\\
\bottomrule
\end{tabular}
\end{table}

\begin{figure}[htbp]
\centering
\begin{minipage}[c]{0.53\linewidth}
\centering
\includefig{width=\linewidth}{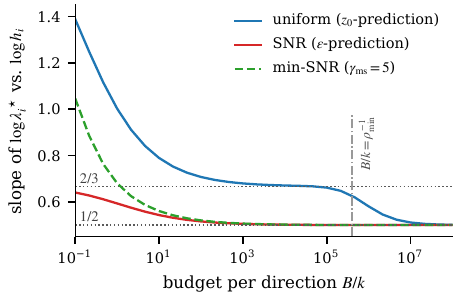}
\end{minipage}\hfill
\begin{minipage}[c]{0.44\linewidth}
\captionof{figure}{Slope of $\log\lambda^\star_i$ against $\log h_i$ for $h_i\propto i^{-1}$ and $k=256$ on the discretized cosine schedule (1,000 steps, $\rho\in[10^{-5.6},10^{8}]$), as a function of the budget per direction. SNR and min-SNR weighting approach the square-root law; uniform weighting stays near $2/3$ until $B/k$ exceeds $\rho_{\min}^{-1}$ (Theorem~\ref{thm:exponent}, Corollary~\ref{cor:cosine}).}
\label{fig:exponent}
\end{minipage}
\end{figure}

\paragraph{Allocation exponent.}
Figure~\ref{fig:exponent} evaluates Theorem~\ref{thm:exponent} on the cosine schedule used to design the reconstruction codes. Under SNR weighting, the slope of $\log\lambda^\star_i$ against $\log h_i$ approaches $1/2$ ($0.505$ at $B/k=10^3$). Under uniform weighting, it remains close to $2/3$ ($0.680$ at $B/k=10^3$ and $0.671$ at $10^4$) and approaches $1/2$ only when $B/k$ exceeds $\rho_{\min}^{-1}\approx4\times10^5$ ($0.501$ at $10^8$), as predicted by Corollary~\ref{cor:cosine}. The exponent is thus a property of the loss weighting rather than of the schedule alone: an allocation designed for a noise-prediction model is more uniform than one designed for a data-prediction model. At $B/k=1$ in this synthetic example, the fitted slope is $0.593$ under SNR weighting. Proposition~\ref{prop:local} accounts for this: over the active directions $\beta_{\mathrm{eff}}$ lies in $[1.62,1.84]$, so the local exponent lies in $[0.545,0.619]$, and the asymptotic value $1/2$ is reached only at far larger budgets. The deployed allocation targets the finite-budget solution of Theorem~\ref{thm:balance}.

\paragraph{Main results.}
Table~\ref{tab:main} reports the mean and standard deviation across seeds 42--44 of the 15-SNR-average differences. All twenty configuration--metric comparisons favor \rbldc{} in every seed. The largest improvements occur for block-local coding on VA-VAE: on the primary configuration, the mean pFID decreases by $17.07$, PSNR increases by $0.855$~dB, SSIM increases by $0.0337$, and LPIPS decreases by $0.0346$; with $b=4$, the corresponding changes are $-22.31$, $+1.303$~dB, $+0.0481$, and $-0.0420$. On the primary configuration, the clean-rFID difference is $+0.00018\pm0.00025$ across the same three seeds. The allocation uses the full measured sensitivity spectrum; the condition number alone does not determine the ordering of observed pFID improvements. Table~\ref{tab:measured-H} in Appendix~\ref{app:results} compares all five spectra and their associated risk and perturbation ratios. These ratios have distinct definitions from the large-budget average-risk gain $\Gamma$ of Proposition~\ref{prop:gain}. A fixed-subset evaluation of the frozen REPA-E tokenizer \citep{leng2025repae} also lowers pFID at all 15 SNRs, by $14.689$ on average, with clean reconstruction parity (Appendix~\ref{app:results}).

\begin{table}[htbp]
\caption{Mean $\pm$ standard deviation across seeds 42--44 of the 15-SNR-average metric differences ($\log_{10}\rho\in[-4,3]$, $\Phi$ equal to the identity). Differences are \rbldc{} minus whitened isotropic coding. Lower is better for pFID and LPIPS; higher is better for PSNR (dB) and SSIM.}
\label{tab:main}
\centering
\footnotesize
\setlength{\tabcolsep}{2.4pt}
\begin{tabular}{@{}lrrrrr@{}}
\toprule
Tokenizer, coding & \multicolumn{1}{c}{$k$} & \multicolumn{1}{c}{$\Delta$pFID} & \multicolumn{1}{c}{$\Delta$PSNR} & \multicolumn{1}{c}{$\Delta$SSIM} & \multicolumn{1}{c@{}}{$\Delta$LPIPS}\\
\midrule
SD-VAE, channel-mix         &   4 & $-1.74\pm0.20$  & $+0.286\pm0.051$ & $+0.0120\pm0.0021$ & $-0.0099\pm0.0033$\\
SD-VAE, block-local         &  16 & $-6.97\pm1.18$  & $+0.700\pm0.063$ & $+0.0292\pm0.0024$ & $-0.0244\pm0.0032$\\
VA-VAE, channel-mix         &  32 & $-2.59\pm1.28$  & $+0.313\pm0.013$ & $+0.0134\pm0.0008$ & $-0.0142\pm0.0019$\\
VA-VAE, block-local         & 128 & $-17.07\pm1.60$ & $+0.855\pm0.040$ & $+0.0337\pm0.0021$ & $-0.0346\pm0.0022$\\
VA-VAE, block-local, $b=4$  & 512 & $-22.31\pm4.57$ & $+1.303\pm0.075$ & $+0.0481\pm0.0035$ & $-0.0420\pm0.0023$\\
\bottomrule
\end{tabular}
\end{table}

\begin{figure}[htbp]
\centering
\includefig{width=\linewidth}{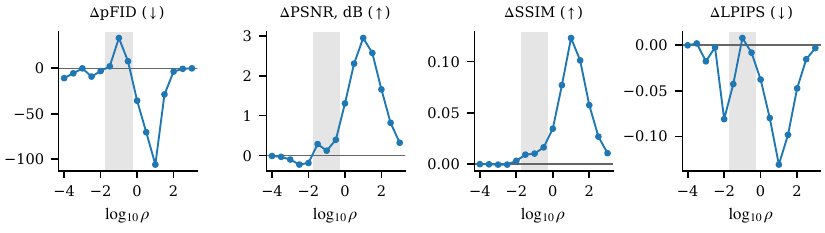}
\caption{Seed-42 per-SNR differences between \rbldc{} and isotropic coding on the primary configuration (VA-VAE, block-local, $k=128$). Arrows indicate the direction of improvement. The shaded range marks the intermediate SNRs at which pFID is higher for \rbldc{} (Section~\ref{sec:exp}, Appendix~\ref{app:intermediate}).}
\label{fig:gaps}
\end{figure}

Figure~\ref{fig:gaps} shows the per-SNR differences for the seed-42 run on the primary configuration. The improvements are concentrated at moderate and high SNR: at $\log_{10}\rho=+1$, pFID decreases by $105.8$, PSNR increases by $2.95$~dB, SSIM increases by $0.123$, and LPIPS decreases by $0.131$. \rbldc{} does not outperform isotropic coding at every SNR: in 14 of the 60 per-SNR comparisons of Figure~\ref{fig:gaps}, isotropic coding yields better scores. Specifically, pFID is higher at $\log_{10}\rho\in\{-1.5,-1.0,-0.5\}$ (by up to $33.2$); PSNR is lower by at most $0.23$~dB for $\log_{10}\rho\le-2$; SSIM is lower by at most $5\times10^{-4}$ for $\log_{10}\rho\le-2.5$; and LPIPS is higher by at most $0.008$ at two SNRs. The median per-SNR pFID difference is $-3.7$, so the average of $-15.4$ is dominated by the SNRs around $\log_{10}\rho=+1$. This run uses one noise realization per SNR; comparisons are paired, and pFID is a set-level statistic over 5,000 images (Appendix~\ref{app:intermediate}).

\paragraph{Sources of the improvement.}
Table~\ref{tab:ablation} decomposes the improvement on the primary configuration into four variants that share the tokenizer, images, noise realizations, and budget: (A) the tokenizer latent with identity coding and no whitening; (B) whitening with isotropic coding; (C) whitening with a dense matrix $G$ trained to minimize an estimate of $\Rk$ under the same budget; and (D) \rbldc{}, which adds the spectral parameterization of Section~\ref{sec:method} and a penalty on violations of the balance condition~\eqref{eq:kkt} (Appendix~\ref{app:impl}). Whitening accounts for the largest single change, which reflects the fact that the analysis is stated in whitened coordinates; we regard it as part of the setup rather than as a contribution. Starting from B, the sensitivity-weighted objective (B$\to$C) and the spectral parameterization with the balance penalty (C$\to$D) contribute almost equally to the average pFID reduction ($7.78$ and $7.62$). Under the same optimizer and budget the dense parameterization does not reach the allocation of Theorem~\ref{thm:balance} (relative spectral error $0.322$) while the spectral one does ($0.005$); both realize the same Gram matrices up to the boundary, so this reflects optimization efficiency, not expressive power. The difference D$-$B reproduces the seed-42 primary result in Table~\ref{tab:seed42} up to rounding. The three feature maps give substantially different eigenvectors of $\hat H$ yet produce pFID curves within approximately 2 points of one another (Appendix~\ref{app:results}).

\begin{table}[htbp]
\caption{Seed-42 component analysis on the primary configuration. Average and $\log_{10}\rho=+1$ values are differences relative to variant A (lower pFID and higher PSNR and SSIM are better). $\Delta_\lambda$ is the relative error of the learned spectrum with respect to the allocation of Theorem~\ref{thm:balance}, and $A_8$ is the alignment of the leading eight-dimensional eigenspaces (1 for identical subspaces). LPIPS is omitted for the reason given in Appendix~\ref{app:ablation}.}
\label{tab:ablation}
\centering
\footnotesize
\setlength{\tabcolsep}{3.0pt}
\begin{tabular}{@{}l rrr rrr rr@{}}
\toprule
& \multicolumn{3}{c}{average over 15 SNRs} & \multicolumn{3}{c}{$\log_{10}\rho=+1$} & & \\
\cmidrule(lr){2-4}\cmidrule(lr){5-7}
Variant & \multicolumn{1}{c}{pFID} & \multicolumn{1}{c}{PSNR} & \multicolumn{1}{c}{SSIM} & \multicolumn{1}{c}{pFID} & \multicolumn{1}{c}{PSNR} & \multicolumn{1}{c}{SSIM} & \multicolumn{1}{c}{$\Delta_\lambda$} & \multicolumn{1}{c@{}}{$A_8$}\\
\midrule
A: identity code, no whitening & $0.00$ & $0.00$ & $0.0000$ & $0.00$ & $0.00$ & $0.0000$ & -- & --\\
B: whitening, isotropic code & $-44.87$ & $+1.22$ & $+0.0401$ & $-94.69$ & $+2.35$ & $+0.1159$ & -- & --\\
C: whitening, dense $G$ for $\Rk$ & $-52.65$ & $+1.66$ & $+0.0576$ & $-159.07$ & $+3.97$ & $+0.1849$ & $0.322$ & $0.698$\\
D: \rbldc{} & $-60.27$ & $+2.04$ & $+0.0714$ & $-200.48$ & $+5.29$ & $+0.2389$ & $0.005$ & $0.99994$\\
\bottomrule
\end{tabular}
\end{table}

In a separate matched evaluation, removing the balance penalty (variant D$^-$) changes the average pFID by only $0.034$, while retaining near-KKT recovery. A matched-budget control that randomly reassigns the KKT spectrum across the eigendirections of $H$ performs worse than isotropic coding, supporting the importance of directional allocation (Appendix~\ref{app:ablation}). A denoising-loss control with 3,000 coding updates gives an average $\Delta$pFID of $+78.27$, compared with $-15.40$ for the primary identity-$\Phi$ code, each relative to the isotropic baseline of its respective evaluation; Appendix~\ref{app:ablation} gives the joint code/DiT training procedure and full results.

\paragraph{Behavior at intermediate SNR.}
At $\log_{10}\rho=-1$ on the primary configuration, PSNR and SSIM favor \rbldc{} while pFID and LPIPS favor isotropic coding. We interpret this as a metric-dependent distortion--perception trade-off, consistent with the framework of \citet{blau2018perception}: minimizing a reconstruction distortion need not improve perceptual quality. Our allocation minimizes a sensitivity-weighted LMMSE surrogate, and its effect on the nonlinear image metrics differs across metrics. The LPIPS result extends the disagreement to a paired VGG-feature distance. Appendix~\ref{app:intermediate} reports the paired comparisons across additional random seeds and discusses the direction-dependent shrinkage in this configuration.

\paragraph{Diffusion-time corruption and latent transmission.}
We evaluate a frozen LightningDiT-XL/1 model \citep{peebles2023dit,yao2025vavae}, trained for 400,000 updates, under added corruption during sampling. At 25\%, 50\%, or 75\% of sampling progress, we pass the current latent state through a Gaussian channel with the learned identity-$\Phi$ code of Table~\ref{tab:ablation}, apply the LMMSE recovery rule, and resume the same sampler. The added channel has $\log_{10}\rho_{\mathrm{ch}}\in\{-1,0,+1,+2\}$; $\rho_{\mathrm{ch}}$ specifies its noise level separately from the diffusion schedule. Each of the 12 settings uses 5,000 generated samples paired across the two codes. FID is lower with \rbldc{} in all 12 settings, with an unweighted average difference of $-25.530$ (Table~\ref{tab:sampling-injection}). Paired CLIP image similarity \citep{radford2021clip} to the corresponding no-injection sample improves in 11 of 12 settings, with an average difference of $+0.01921$.

\begin{table}[htbp]
\caption{FID after a single corruption during sampling with the frozen 400k-update LightningDiT-XL/1 model and the learned identity-$\Phi$ code ($k=B=128$). Each entry is isotropic / \rbldc{} for 5,000 samples. The matched no-injection FID is $9.612$. Progress is the fraction of sampling intervals completed; columns specify the added channel SNR.}
\label{tab:sampling-injection}
\centering
\small
\setlength{\tabcolsep}{5pt}
\begin{tabular}{@{}lrrrr@{}}
\toprule
& \multicolumn{4}{c}{$\log_{10}\rho_{\mathrm{ch}}$}\\
\cmidrule(lr){2-5}
Progress & \multicolumn{1}{c}{$-1$} & \multicolumn{1}{c}{$0$} & \multicolumn{1}{c}{$+1$} & \multicolumn{1}{c@{}}{$+2$}\\
\midrule
25\% & $241.938$ / $191.122$ & $121.488$ / $106.438$ & $14.056$ / $13.395$ & $9.853$ / $9.850$\\
50\% & $323.545$ / $235.477$ & $147.836$ / $124.612$ & $13.438$ / $12.440$ & $9.770$ / $9.710$\\
75\% & $335.773$ / $246.106$ & $179.753$ / $144.848$ & $14.120$ / $11.337$ & $9.804$ / $9.676$\\
\bottomrule
\end{tabular}
\end{table}

In a separate transmission experiment, we re-encode 10,000 clean samples from the fixed generator with VA-VAE, transmit their coded latents through the added Gaussian channel, and recover and decode them. This evaluation uses an analytic code with a Gram eigenvalue ratio capped at 4 and a weighted-Procrustes orientation. Over the 15-SNR grid, sample FID improves at 13 SNRs, with an average difference of $-7.23$, and paired CLIP similarity improves at 14 SNRs. At $\log_{10}\rho_{\mathrm{ch}}=+1.5$ and $+2$, FID changes from $46.11$ to $31.91$ and from $13.64$ to $11.33$, respectively. The generator remains frozen in both evaluations. Appendix~\ref{app:results} gives the distinct code variants, complete results, and evaluation protocols.

\section{Limitations and conclusion}
\label{sec:conclusion}

The theory measures the distortion between the clean and the noisy decoded images, while the reconstruction study compares the noisy reconstruction with the original; the two differ by a cross term that depends on the allocation even when the clean path is fixed (Appendix~\ref{app:nonlinear}). The risk is a surrogate in two further respects: it is the error of the LMMSE estimator, so a trained denoiser adds the terms $\mathcal E_\theta$ and $\mathcal A$ of Proposition~\ref{prop:excess}, which we do not measure; and its link to nonlinear feature distortion is asymptotically tight under noise-prediction weighting but not under uniform weighting, where an explicit example has a different optimum (Appendix~\ref{app:nonlinear}). The main reconstruction experiments use one weighting and one budget ($B=k$), and evaluate image metrics. The operator-norm bound inferred from the Frobenius whitening residual can be loose (Remark~\ref{rem:colored}). The layer is linear and block-local and the tokenizer is frozen; joint training of the tokenizer and coding layer, nonlinear codes, and input-dependent allocations are natural extensions.

An invertible linear layer between a tokenizer and a diffusion process is a per-direction shift of the noise schedule. Allocating a fixed latent energy by decoder sensitivity has a unique solution, computed by a deterministic spectral solver, whose large-budget form is a power law with an exponent set by the loss weighting. Our experiments show improved reconstruction from noisy latents on two tokenizers and five configurations without changing clean reconstruction. Further tests with a frozen diffusion model show lower average FID under added corruption during sampling and after generation.

\newpage
\subsection*{Reproducibility statement}
The mathematical assumptions and complete proofs are given in Sections~\ref{sec:setup}--\ref{sec:theory} and Appendices~\ref{app:prelim}--\ref{app:block}. Algorithm~\ref{alg:rbldc} and Appendix~\ref{app:impl} specify the spectral solver and trained parameterization. Section~\ref{sec:exp} and Appendices~\ref{app:results} and~\ref{app:details} describe the datasets, tokenizers, coding budgets, SNR protocols, metrics, and repeated-run comparisons. Appendix~\ref{app:results} reports measured sensitivity statistics and the fixed-subset REPA-E evaluation results.

\subsection*{Ethics statement}
This work studies a component of image tokenizers using a public dataset and public pretrained models. It does not involve human subjects or personal data. More robust latent spaces could increase the fidelity of generative image models, which carries the general dual-use risks of such models.

\subsection*{AI use statement}
In this work, we used generative AI tools to assist with checking and writing mathematical proofs, with feedback on the experimental methodology, with literature search and the verification of bibliographic information, with writing code for the numerical evaluation of the theoretical results (Figure~\ref{fig:exponent}) and for plotting reported values (Figure~\ref{fig:gaps}). We did not use generative AI tools to produce experimental measurements. The authors verified all mathematical statements and proofs, checked the AI-assisted code for correctness, and checked all references against their original sources. We take responsibility for the final content of this work, including text, claims, and artifacts produced with the aid of generative AI.

\begin{ack}
This work was supported by the Institute of Information \& Communications
Technology Planning \& Evaluation (IITP) grant funded by the Korea government
(MSIT) (RS-2026-25530181, Development of Inter-Domain Quantum Security System
Combining PQC and QKD, 50\%), and by the InnoCORE program of the Ministry of
Science and ICT (26-InnoCORE-01, 50\%). The authors declare no competing interests.
\end{ack}

\bibliography{mybib}
\bibliographystyle{plainnat}

\appendix
\section{Related work}
\label{sec:related}

\paragraph{Tokenizers for latent diffusion.}
LDMs rely on pretrained autoencoders \citep{rombach2022ldm}, and recent work aligns tokenizer latents with pretrained representations \citep{yao2025vavae,page2026sdvae}, trains the tokenizer jointly with the diffusion model \citep{leng2025repae}, or replaces the encoder with a representation model and a trained decoder \citep{zheng2026rae,tong2026raet2i}. \citet{leng2025repae} report that end-to-end training with the diffusion loss alone reduces latent variance and degrades generation, which motivates fixing the latent energy when a coding layer is optimized. Robustness of the latent space to perturbations has been measured and improved by post-training \citep{qiu2025robusttok} and by a variance expansion loss \citep{li2026veloss}, and adversarial robustness has been analyzed for fixed pipelines \citep{zhang2023robustldm}. \rbldc{} keeps the tokenizer fixed and redistributes a fixed latent energy across directions. The decoder post-training of \citet{qiu2025robusttok} covers both discrete and continuous tokenizers and uses generator-dependent inputs; because it adapts the decoder under generator-specific perturbations, it addresses a setting orthogonal to our fixed-tokenizer, matched-trace Gaussian channel and LMMSE recovery protocol.

\paragraph{Noise schedules and loss weighting.}
The cosine schedule \citep{nichol2021improved}, the SNR parameterization of diffusion \citep{kingma2021vdm}, and the reading of diffusion objectives as weighted evidence lower bounds \citep{kingma2023elbo} underlie our treatment of $w$; min-SNR weighting balances the contributions of noise levels \citep{hang2023minsnr}. Scaling the input or shifting the schedule changes the SNR of all directions equally \citep{chen2023noise,hoogeboom2023simple,esser2024sd3,zheng2026rae}, and component-wise schedules in pixel, principal-component, Fourier and wavelet bases \citep{gerdes2024gud,falck2025fourier} choose per-component noise levels from the data representation. \rbldc{} chooses a per-direction shift from decoder sensitivity under an energy budget, and relates its shape to the loss weighting.

\paragraph{Joint source--channel coding and linear precoding.}
Source and channel coding are separable in the asymptotic regime \citep{shannon1948}, and learned joint source--channel codes can outperform separate designs at finite block lengths and low SNR \citep{bourtsoulatze2019deepjscc,choi2019neuraljscc,lee2023isec}. Optimal linear coding of vector sources over Gaussian channels \citep{leepetersen1976} and the design of linear transceivers under weighted mean-square-error criteria \citep{scaglione2002,palomar2003,xing2013weighted} contain the single-SNR form of Theorems~\ref{thm:align} and~\ref{thm:balance}, and linear transmission over a fading channel with the channel distribution averaged out gives the corresponding integrated allocation \citep{tan2016linear}. Diffusion models have been used at the receiver of semantic and joint source--channel communication systems \citep{qiao2024gensemcom,yang2024diffjscc} and in low-rate image compression \citep{xia2025diffpc,ke2025resulic}. We apply this viewpoint to the interface between a tokenizer and a diffusion model, where the channel is the forward process and the averaging measure is the loss weighting.

\paragraph{Information-theoretic views of diffusion.}
The I-MMSE relation connects mutual information and estimation error in Gaussian channels \citep{guo2005immse}, information-theoretic diffusion relates the data distribution to optimal denoising \citep{kong2023itdiffusion}, and U-Nets admit a belief-propagation interpretation \citep{mei2024unetbp}. The risk $\Rk$ is an estimation error integrated over SNR in the same spirit.

\section{Assumptions and matrix lemmas}
\label{app:prelim}

Throughout the appendix we use the following standing assumptions. Norms of vectors are Euclidean, $\|\cdot\|_{\mathrm{op}}$ is the operator norm, and $\preceq$ is the positive semidefinite (PSD) order.

\begin{assumption}[Latent]\label{ass:latent}
$S\in\R^k$ satisfies $\E S=0$ and $\E SS^\top=I_k$.
\end{assumption}

\begin{assumption}[Code and channel]\label{ass:channel}
$G\in\R^{m\times k}$ with $m\ge k$, $M=G^\top G$, and $Y=\sqrt\rho\,GS+\varepsilon$ with $\varepsilon\sim\N(0,I_m)$ independent of $S$. The budget is $\tr M\le B$.
\end{assumption}

\begin{assumption}[SNR measure]\label{ass:measure}
$\nu$ is a nonzero finite Borel measure on $(0,\rho_{\max}]$ for some $\rho_{\max}<\infty$. The measure of Section~\ref{sec:theory}, supported in $[\rho_{\min},\rho_{\max}]$, is a special case.
\end{assumption}

\begin{assumption}[Sensitivity]\label{ass:sens}
$H=\E[J_F(S)^\top J_F(S)]$ is finite and $H=U\diag(h_1,\dots,h_k)U^\top$ with $h_1\ge\dots\ge h_k>0$. (If $H$ is singular, all statements apply on its support.)
\end{assumption}

\begin{lemma}[Kernels]
\label{lem:kernels}
Under Assumption~\ref{ass:measure}, $\varphi$ and $m=-\varphi'$ are infinitely differentiable on $[0,\infty)$; $\varphi$ is strictly decreasing and strictly convex; $m$ is positive and strictly decreasing with $m(0)=\int\rho\dnu<\infty$; $\varphi(\lambda)\to0$ and $m(\lambda)\to0$ as $\lambda\to\infty$; and, with $c^{-1}:=\int\rho^{-1}\dnu\in(0,\infty]$,
\[
\lambda\,\varphi(\lambda)\uparrow c^{-1}\qquad\text{and}\qquad \lambda^2m(\lambda)\uparrow c^{-1}\qquad(\lambda\to\infty).
\]
\end{lemma}
\begin{proof}
For $\lambda\ge0$ and $\rho\in(0,\rho_{\max}]$, the $j$-th derivative of $(1+\rho\lambda)^{-1}$ with respect to $\lambda$ is bounded in absolute value by $j!\,\rho_{\max}^j$, which is $\nu$-integrable because $\nu$ is finite. Differentiation under the integral is therefore justified, and
$\varphi'(\lambda)=-\int\rho(1+\rho\lambda)^{-2}\dnu=-m(\lambda)<0$, $\varphi''(\lambda)=2\int\rho^2(1+\rho\lambda)^{-3}\dnu>0$, and $m'=-\varphi''<0$. The limits $\varphi,m\to0$ follow from dominated convergence. Finally, $\lambda/(1+\rho\lambda)\uparrow\rho^{-1}$ and $\lambda^2\rho/(1+\rho\lambda)^2=\rho^{-1}\big(\rho\lambda/(1+\rho\lambda)\big)^2\uparrow\rho^{-1}$ monotonically in $\lambda$, so the last two limits follow from monotone convergence.
\end{proof}

\begin{lemma}[Realizable Gram matrices]
\label{lem:realize}
For every $m\ge k$, $\{G^\top G:G\in\R^{m\times k},\ \tr G^\top G\le B\}=\{M\in\R^{k\times k}:M\succeq0,\ \tr M\le B\}$.
\end{lemma}
\begin{proof}
$G^\top G\succeq0$ and $\tr G^\top G=\tr M$. Conversely, $G=\begin{psmallmatrix}M^{1/2}\\0\end{psmallmatrix}\in\R^{m\times k}$ satisfies $G^\top G=M$.
\end{proof}

\begin{lemma}[Operator convexity]
\label{lem:opconvex}
For $\rho>0$, the map $X\mapsto(I+\rho X)^{-1}$ is operator convex on the PSD cone: for $X,Y\succeq0$ and $t\in[0,1]$,
$(I+\rho(tX+(1-t)Y))^{-1}\preceq t(I+\rho X)^{-1}+(1-t)(I+\rho Y)^{-1}$. Consequently, $X\mapsto\varphi(X)=\int(I+\rho X)^{-1}\dnu$ is operator convex, and $\varphi\big(\sum_jp_jX_j\big)\preceq\sum_jp_j\varphi(X_j)$ for finitely many $X_j\succeq0$ and probability weights $p_j$.
\end{lemma}
\begin{proof}
The operator convexity of $X\mapsto X^{-1}$ on positive definite matrices is classical \citep[Chapter~V]{bhatia1997matrix}; we include a short proof. It uses the fact that a symmetric block matrix whose upper-left block is positive definite is PSD if and only if the Schur complement of that block is PSD \citep{horn2012matrix}. For $P\succ0$, this gives
\begin{equation*}
\begin{pmatrix}
P&I\\
I&P^{-1}
\end{pmatrix}\succeq0,
\end{equation*}
because its Schur complement is $P^{-1}-P^{-1}=0$.
Let $A=I+\rho X\succ0$ and $C=I+\rho Y\succ0$. Taking a convex combination yields
\begin{equation*}
\begin{pmatrix}
tA+(1-t)C&I\\
I&tA^{-1}+(1-t)C^{-1}
\end{pmatrix}\succeq0.
\end{equation*}
Since $tA+(1-t)C\succ0$, its Schur complement satisfies
\begin{equation*}
tA^{-1}+(1-t)C^{-1}
-\big(tA+(1-t)C\big)^{-1}\succeq0.
\end{equation*}
As $tA+(1-t)C=I+\rho(tX+(1-t)Y)$, this is the first claim.
Integrating against the nonnegative measure $\nu$ preserves the PSD order, which gives operator convexity of $\varphi$; this agrees with the spectral definition because $\int(I+\rho X)^{-1}\dnu=U_X\diag\big(\varphi(\lambda_i(X))\big)U_X^\top$.
The finite version follows by induction on the number of terms: if $p_n<1$, write $\sum_jp_jX_j=p_nX_n+(1-p_n)\sum_{j<n}\frac{p_j}{1-p_n}X_j$ and apply the two-point inequality followed by the induction hypothesis.
\end{proof}

\begin{lemma}[Pinching]
\label{lem:pinch}
Let $\R^k=\bigoplus_{b=1}^LE_b$ be an orthogonal decomposition with orthogonal projections $\Pi_b$, and let $\pin(X)=\sum_b\Pi_bX\Pi_b$. For every $X\succeq0$ and every symmetric $Y$: $\varphi(\pin(X))\preceq\pin(\varphi(X))$, $\tr\pin(X)=\tr X$, and $\tr(Y\,\pin(X))=\tr(\pin(Y)\,X)$.
\end{lemma}
\begin{proof}
For $\epsilon\in\{-1,1\}^L$, let $D_\epsilon=\sum_b\epsilon_b\Pi_b$, a symmetric orthogonal matrix. If $\epsilon$ is uniform on $\{-1,1\}^L$, then $\E[\epsilon_b\epsilon_c]=\mathbf{1}[b=c]$, so $\E[D_\epsilon XD_\epsilon]=\sum_{b,c}\E[\epsilon_b\epsilon_c]\Pi_bX\Pi_c=\pin(X)$. By Lemma~\ref{lem:opconvex} with equal weights $2^{-L}$, and because matrix functions commute with orthogonal similarity,
$\varphi(\pin(X))\preceq2^{-L}\sum_\epsilon\varphi(D_\epsilon XD_\epsilon)=2^{-L}\sum_\epsilon D_\epsilon\varphi(X)D_\epsilon=\pin(\varphi(X))$. The trace identities follow from $\Pi_b^2=\Pi_b$ and the cyclicity of the trace.
\end{proof}

\begin{lemma}[Eigenvalue trace inequality]
\label{lem:trace}
Let $A,B\succeq0$ have eigenvalues $a_1\ge\dots\ge a_k$ and $b_1\ge\dots\ge b_k$. Then $\sum_ia_ib_{k+1-i}\le\tr(AB)\le\sum_ia_ib_i$.
\end{lemma}
\begin{proof}
The upper bound is von Neumann's trace inequality \citep{mirsky1975trace}, since singular values and eigenvalues coincide for PSD matrices. For the lower bound, $B'=b_1I-B\succeq0$ has eigenvalues $b_1-b_{k+1-i}$ in decreasing order, so the upper bound applied to $A$ and $B'$ gives
$\tr(AB)=b_1\tr A-\tr(AB')\ge b_1\sum_ia_i-\sum_ia_i(b_1-b_{k+1-i})=\sum_ia_ib_{k+1-i}$.
\end{proof}

\section{Proofs and Extensions for Section~\ref{sec:setup}}
\label{app:proofs-setup}

\begin{proof}[Proof of Proposition~\ref{prop:lmmse}]
(a) Under Assumptions~\ref{ass:latent} and~\ref{ass:channel}, $\cov(S,Y)=\sqrt\rho\,G^\top$ and $\cov(Y)=\rho GG^\top+I_m\succ0$. The linear estimator with the smallest error covariance is $WY$, where
\begin{equation*}
\begin{aligned}
W
&=\cov(S,Y)\cov(Y)^{-1}\\
&=\sqrt\rho\,G^\top(\rho GG^\top+I_m)^{-1}\\
&=\sqrt\rho\,(I_k+\rho G^\top G)^{-1}G^\top.
\end{aligned}
\end{equation*}
The last equality is the push-through identity. By the Woodbury identity, its error covariance is
\begin{equation*}
\begin{aligned}
\Sigma_\rho(M)
&=I_k-\rho G^\top(\rho GG^\top+I_m)^{-1}G\\
&=(I_k+\rho M)^{-1}.
\end{aligned}
\end{equation*}
For any affine estimator $AY+c$, the error second-moment matrix equals $(I_k+\rho M)^{-1}+(A-W)\cov(Y)(A-W)^\top+cc^\top\succeq(I_k+\rho M)^{-1}$, because $\E[(S-WY)Y^\top]=0$ and $\E S=\E Y=0$.

(b) If $S$ is Gaussian, then $(S,Y)$ is jointly Gaussian, $\E[S\mid Y]$ is linear in $Y$ and equals $WY$, and the error $S-WY$ is Gaussian and independent of $Y$ with covariance $(I_k+\rho M)^{-1}$. For general $S$, the tower property gives $\E[(S-f(Y))(S-f(Y))^\top]=\Sigma_{\mathrm{B}}+\E[(\E[S\mid Y]-f(Y))(\E[S\mid Y]-f(Y))^\top]\succeq\Sigma_{\mathrm{B}}$ for every measurable $f$, where $\Sigma_{\mathrm{B}}$ is the error covariance of $\E[S\mid Y]$; taking $f(Y)=WY$ yields $\Sigma_{\mathrm{B}}\preceq(I_k+\rho M)^{-1}$.
\end{proof}

\begin{proof}[Proof of Proposition~\ref{prop:spectrum}]
By Proposition~\ref{prop:lmmse}, $\Sigma_\rho$ is a function of $M$, and $\Rk(M)=\int\tr(H\Sigma_\rho(M))\dnu$. By Lemma~\ref{lem:realize}, the set of attainable $M$ under the budget does not depend on $m\ge k$, so neither does the minimal risk. For an orthogonal $Q\in\R^{m\times m}$, $(QG)^\top(QG)=G^\top G$. If the channel noise has covariance $C\succ0$ instead of $I_m$, whitening $Y$ by $C^{-1/2}$ gives the covariance and risk formulas with $G^\top C^{-1}G$ in place of $M$. The feasible generalized Gram matrices under $\tr(G^\top G)\le B$ depend on $C$, so the dimension-invariance argument above uses isotropic channel noise.
\end{proof}

\begin{proof}[Proof of Lemma~\ref{lem:loss}]
With $z_t=\alpha_tGs+\sigma_t\varepsilon$, define the estimates induced by $\hat s$ as
\begin{equation*}
\hat\varepsilon=\frac{z_t-\alpha_tG\hat s}{\sigma_t},
\qquad
\hat z_0=G\hat s,
\qquad
\hat v=\alpha_t\hat\varepsilon-\sigma_t\hat z_0.
\end{equation*}
Because $G$ is invertible, these parameterizations are in one-to-one correspondence. Then $\varepsilon-\hat\varepsilon=(\alpha_t/\sigma_t)G(\hat s-s)$, $z_0-\hat z_0=G(s-\hat s)$, and $v-\hat v=\alpha_t(\varepsilon-\hat\varepsilon)-\sigma_t(z_0-\hat z_0)=\big(\alpha_t^2/\sigma_t+\sigma_t\big)G(\hat s-s)=\sigma_t^{-1}G(\hat s-s)$, using $\alpha_t^2+\sigma_t^2=1$. The three identities follow from $\alpha_t^2/\sigma_t^2=\rho_t$ and $\sigma_t^{-2}=1+\rho_t$. With $t$ uniform, the expected loss is $\int w(t)\,\E[(\hat s-s)^\top M(\hat s-s)]\,dt$ with the stated $w$.
\end{proof}

\begin{proposition}[Imperfect whitening]
\label{prop:colored}
Let $\E S=0$ and $\E SS^\top=C\succ0$, and let $Y=\sqrt\rho\,GS+\varepsilon$ as in Assumption~\ref{ass:channel}.

(a) The LMMSE error covariance of $S$ is $(C^{-1}+\rho M)^{-1}=C^{1/2}(I_k+\rho\tilde M)^{-1}C^{1/2}$ with $\tilde M=C^{1/2}MC^{1/2}$, the coded energy is $\E\|GS\|^2=\tr(GCG^\top)=\tr\tilde M$, and $\tr\big(H(C^{-1}+\rho M)^{-1}\big)=\tr\big(\tilde H(I_k+\rho\tilde M)^{-1}\big)$ with $\tilde H=C^{1/2}HC^{1/2}$. All results of Sections~\ref{sec:setup}--\ref{sec:method} therefore apply verbatim to $(\tilde G,\tilde H)$ with $\tilde G=GC^{1/2}$.

(b) The estimator of Proposition~\ref{prop:lmmse}, which is built for a white latent, has error covariance $A_\rho(C+\rho M)A_\rho=A_\rho+A_\rho(C-I_k)A_\rho$ with $A_\rho=(I_k+\rho M)^{-1}$; this differs from $(C^{-1}+\rho M)^{-1}$ in general.

(c) If $\|C-I_k\|_{\mathrm{op}}\le\tau<1$, then for every $M\succeq0$ both the $C$-aware risk $\Rk^C(M)=\int\tr(H(C^{-1}+\rho M)^{-1})\dnu$ and the risk $\Rk^{\mathrm{wf}}(M)$ of the white-model estimator lie in $[(1-\tau)\Rk(M),(1+\tau)\Rk(M)]$, where $\Rk$ is the white-model risk. Consequently the minimizer $M^{\mathrm w}$ of $\Rk$ over a feasible set satisfies $\Rk^\bullet(M^{\mathrm w})\le\frac{1+\tau}{1-\tau}\min_M\Rk^\bullet(M)$ for $\bullet\in\{C,\mathrm{wf}\}$.

(d) Under the change of variables of (a), the physical energy constraint $\E\|GS\|^2=\tr(MC)\le B$ becomes $\tr\tilde M\le B$, whereas the algebraic constraint $\tr M\le B$ becomes $\tr(C^{-1}\tilde M)\le B$; the two agree only for $C=I_k$ (for $k=1$, $C=2$ and $M=1$ one has $\tr M=1$ but $\tr\tilde M=2$). The floor $M\succeq\ell I_k$ becomes $\tilde M\succeq\ell C$, and a block-diagonal $M$ has block-diagonal $\tilde M$ if and only if $C^{1/2}$ preserves the partition, for instance when $C$ is block-diagonal for it. We impose the budget and the floor in the whitened coordinates, where they read $\tr\tilde M\le B$ and $\tilde M\succeq\ell I_k$.
\end{proposition}

\begin{proof}
(a) The first identity follows from
\begin{equation*}
\begin{aligned}
C^{1/2}(I_k+\rho C^{1/2}MC^{1/2})^{-1}C^{1/2}
&=\big(C^{-1/2}(I_k+\rho C^{1/2}MC^{1/2})C^{-1/2}\big)^{-1}\\
&=(C^{-1}+\rho M)^{-1}.
\end{aligned}
\end{equation*}
The LMMSE formula itself is Proposition~\ref{prop:lmmse} applied to the whitened variable $\tilde S=C^{-1/2}S$, for which $GS=\tilde G\tilde S$. The energy and trace identities follow from the cyclicity of the trace.

(b) With $W=\sqrt\rho\,A_\rho G^\top$ and $Y=\sqrt\rho\,GS+\varepsilon$, we have
\begin{equation*}
\begin{aligned}
S-WY
&=(I_k-\rho A_\rho M)S-\sqrt\rho\,A_\rho G^\top\varepsilon\\
&=A_\rho S-\sqrt\rho\,A_\rho G^\top\varepsilon,
\end{aligned}
\end{equation*}
because $I_k-\rho A_\rho M=A_\rho$. Taking covariances gives
\begin{equation*}
\begin{aligned}
\cov(S-WY)
&=A_\rho CA_\rho+\rho A_\rho MA_\rho\\
&=A_\rho(C+\rho M)A_\rho\\
&=A_\rho+A_\rho(C-I_k)A_\rho.
\end{aligned}
\end{equation*}
The last equality uses $A_\rho(I_k+\rho M)A_\rho=A_\rho$.

(c) From $C^{-1}\succeq(1+\tau)^{-1}I_k$ and $\rho M\succeq(1+\tau)^{-1}\rho M$ we get $C^{-1}+\rho M\succeq(1+\tau)^{-1}(I_k+\rho M)$, hence $(C^{-1}+\rho M)^{-1}\preceq(1+\tau)A_\rho$; the lower bound follows in the same way from $C^{-1}\preceq(1-\tau)^{-1}I_k$. For the white-model estimator, $\|C-I_k\|_{\mathrm{op}}\le\tau$ gives $-\tau A_\rho^2\preceq A_\rho(C-I_k)A_\rho\preceq\tau A_\rho^2$, and $0\preceq A_\rho\preceq I_k$ gives $A_\rho^2\preceq A_\rho$, so the covariance lies in $[(1-\tau)A_\rho,(1+\tau)A_\rho]$. Multiplying by $H\succeq0$, taking traces and integrating gives both two-sided bounds, and the suboptimality factor follows as in the proof of Proposition~\ref{prop:sandwich}.

(d) The first claim follows from $\tr\tilde M=\tr(C^{1/2}MC^{1/2})=\tr(MC)$ and the second from $M=C^{-1/2}\tilde MC^{-1/2}$, so that $\tr M=\tr(C^{-1}\tilde M)$. The floor statement follows from $\tilde M-\ell C=C^{1/2}(M-\ell I_k)C^{1/2}$, and the last claim is immediate.
\end{proof}

\begin{proof}[Proof of Proposition~\ref{prop:excess}]
Write $\Delta=\Shat_{\mathrm B}-\Shat_\theta$, a function of $Y$.
Since $\E[S-\Shat_{\mathrm B}\mid Y]=0$, the tower property gives $\E\big[(S-\Shat_{\mathrm B})^\top H\Delta\big]=0$.
Thus, for each $\rho$,
\begin{equation*}
\E\|H^{1/2}(S-\Shat_\theta)\|^2
=\E\|H^{1/2}(S-\Shat_{\mathrm B})\|^2
 +\E\|H^{1/2}\Delta\|^2.
\end{equation*}
Integrating against $\nu$ gives $\mathcal L_\theta(M)=\Rk_{\mathrm B}(M)+\mathcal E_\theta(M)$, which is~\eqref{eq:excess} because $\Rk_{\mathrm B}=\Rk-\mathcal A$.
The gap satisfies $\mathcal A(M)\ge0$ by Proposition~\ref{prop:lmmse}(b), with equality for Gaussian $S$.
The comparison rule is a rearrangement of $\mathcal L_\theta(M)<\mathcal L_\theta(M')$.
\end{proof}

\begin{example}[The Gaussian form of the comparison rule fails in general]
\label{ex:bayesgap}
Let $S_1\sim\N(0,1)$ and $S_2$ be an independent symmetric sign, so that $\E SS^\top=I_2$, and let $H=I_2$, $\nu=\delta_1$, $M=\diag(4,1)$ and $M'=\diag(1,4)$. Both codes have the same LMMSE risk, $\Rk(M)=\Rk(M')=\tfrac15+\tfrac12=0.7$, but different Bayes risks, $\Rk_{\mathrm B}(M)=0.6496$ and $\Rk_{\mathrm B}(M')=0.5686$, because the sign component is estimated much more accurately at the higher SNR. For the estimator $f\equiv0$ both actual risks equal $\E\|S\|^2=2$, yet $\mathcal E_\theta(M)-\mathcal E_\theta(M')=\Rk_{\mathrm B}(M')-\Rk_{\mathrm B}(M)=-0.0810<0=\Rk(M')-\Rk(M)$. The rule that omits $\mathcal A$ would declare $M$ strictly better, whereas the rule of Proposition~\ref{prop:excess}, in which $\mathcal A(M)-\mathcal A(M')=-0.0810$ cancels the same quantity, correctly declares the two codes equivalent for this estimator.
\end{example}

\paragraph{On Remark~\ref{rem:scope}.}
If the score of the marginal distribution of $Z_t$ is known exactly and the reverse-time process is integrated exactly, the reverse process reproduces the forward marginals, so the generated $Z_0$ has the law of $GS$ and $G^{-1}Z_0$ has the law of $S$, whatever the invertible $G$. Without external corruption, a coding layer can therefore affect generation only through the errors of a trained model, which our analysis does not model.

\section{Proofs and Extensions for Section~\ref{sec:theory}}
\label{app:proofs-theory}

\begin{proof}[Proof of Theorem~\ref{thm:align}]
Let $M\succeq0$ have eigenvalues $\lambda_1\ge\dots\ge\lambda_k$. The matrix $\varphi(M)\succeq0$ has eigenvalues $\varphi(\lambda_1)\le\dots\le\varphi(\lambda_k)$, because $\varphi$ is decreasing (Lemma~\ref{lem:kernels}); in decreasing order they are $\varphi(\lambda_{k+1-i})$. The lower bound of Lemma~\ref{lem:trace} applied to $A=H$ and $B=\varphi(M)$ gives $\Rk(M)=\tr(H\varphi(M))\ge\sum_ih_i\varphi(\lambda_i)$. If $M=U\diag(\lambda)U^\top$, then $\varphi(M)=U\diag(\varphi(\lambda))U^\top$ and $\tr(H\varphi(M))=\sum_ih_i\varphi(\lambda_i)$. Both constraints, $\tr M=\sum_i\lambda_i\le B$ and $M\succeq\ell I_k$, that is $\lambda_i\ge\ell$, depend only on the spectrum, so replacing any feasible $M$ by the aligned matrix with the same spectrum preserves feasibility and does not increase the risk. Conversely, every feasible $\lambda$ in \eqref{eq:scalar} is the spectrum of the feasible aligned matrix $U\diag(\lambda)U^\top$. Hence the two problems have the same optimal value, and an optimal $M$ is obtained by aligning an optimal $\lambda$.
\end{proof}

\begin{proof}[Proof of Theorem~\ref{thm:balance}]
The feasible set $\{\lambda:\lambda_i\ge\ell,\ \sum_i\lambda_i\le B\}$ is nonempty (because $B>k\ell$), compact and convex, and the objective is continuous and strictly convex because each $h_i\varphi$ is strictly convex (Lemma~\ref{lem:kernels}) and $h_i>0$.
Hence a unique minimizer $\lambda^\star$ exists.
If $\sum_i\lambda_i^\star<B$, increasing any coordinate would strictly decrease the objective because $\varphi$ is strictly decreasing; hence the budget is exhausted.
The constraints are affine, so the Karush--Kuhn--Tucker (KKT) conditions are necessary and sufficient: there are $\mu\ge0$ and $\eta_i\ge0$ such that
\begin{equation*}
\begin{aligned}
\eta_i(\lambda_i^\star-\ell)&=0,\\
h_i\varphi'(\lambda_i^\star)+\mu-\eta_i&=0.
\end{aligned}
\end{equation*}
Since $\varphi'=-m$, these are the conditions in~\eqref{eq:kkt}, and $\mu=h_im(\lambda_i^\star)>0$ whenever $\lambda_i^\star>\ell$.
Because $m$ is strictly decreasing, the allocation at multiplier $\mu$ is
\begin{equation*}
\lambda_i(\mu)=
\begin{cases}
m^{-1}(\mu/h_i),&h_im(\ell)>\mu,\\
\ell,&h_im(\ell)\le\mu.
\end{cases}
\end{equation*}
The sum $\sum_i\lambda_i(\mu)$ is continuous and strictly decreasing in $\mu$ wherever it exceeds $k\ell$, and Algorithm~\ref{alg:rbldc} finds the $\mu$ with $\sum_i\lambda_i(\mu)=B$.

For monotonicity, suppose $h_i\ge h_j$ and $\lambda_i^\star<\lambda_j^\star$. Then $\lambda_j^\star>\ell$. If $\lambda_i^\star>\ell$ as well, then $h_im(\lambda_i^\star)>h_jm(\lambda_j^\star)$ because $m$ is strictly decreasing, contradicting the equality of both sides with $\mu$. If $\lambda_i^\star=\ell$, then $h_im(\ell)\le\mu=h_jm(\lambda_j^\star)<h_jm(\ell)\le h_im(\ell)$, a contradiction. Finally, if $\mu$ did not tend to zero as $B\to\infty$, every $\lambda_i^\star$ would remain bounded (because $m(\lambda)\to0$), contradicting $\sum_i\lambda_i^\star=B$; hence $\mu\to0$ and eventually $h_im(\ell)>\mu$ for all $i$, so $\lambda_i^\star>\ell$ for every $i$ and the floor is inactive.
\end{proof}

\begin{example}[Without a floor the optimum can be singular]
\label{ex:singular}
Let $H=\diag(100,1)$, $\nu=\delta_1$ and $B=2$, so that $\varphi(\lambda)=(1+\lambda)^{-1}$ and $m(\lambda)=(1+\lambda)^{-2}$. Problem~\eqref{eq:scalar} with $\ell=0$ minimizes $100/(1+\lambda_1)+1/(1+\lambda_2)$ subject to $\lambda_1+\lambda_2\le2$. An interior stationary point would require $1+\lambda_1=10(1+\lambda_2)$, which is infeasible, so the solution is $\lambda^\star=(2,0)$, with $\mu=h_1m(2)=100/9$ and $h_2m(0)=1\le\mu$, in agreement with~\eqref{eq:kkt}. The corresponding code is singular. With a floor $\ell>0$ the same instance gives $\lambda^\star=(2-\ell,\ell)$ whenever $100/(3-\ell)^2\ge1/(1+\ell)^2$, which holds for every feasible floor $\ell<1$, and the code is then invertible with $\mathrm{cond}(G)=\sqrt{(2-\ell)/\ell}$.
\end{example}

The next proposition gives the behavior of $m$ at large arguments.

\begin{proposition}[Kernel asymptotics]
\label{prop:asymp}
Let $\nu$ be finite on $(0,\rho_{\max}]$ with $\nu(d\rho)=C\rho^{\gamma-1}(1+o(1))\,d\rho$ as $\rho\to0$, for some $C,\gamma>0$. As $\lambda\to\infty$,
\[
m(\lambda)\sim\begin{cases}
C\,\dfrac{\gamma\pi}{\sin(\gamma\pi)}\,\lambda^{-(1+\gamma)}, & 0<\gamma<1,\\[6pt]
C\,\lambda^{-2}\log\lambda, & \gamma=1,\\[2pt]
c^{-1}\,\lambda^{-2}, & \gamma>1,
\end{cases}
\]
where $c^{-1}=\int\rho^{-1}\dnu$ is finite exactly when $\gamma>1$.
\end{proposition}
\begin{proof}
Let $\epsilon(\rho)$ be defined by $\nu(d\rho)=C\rho^{\gamma-1}(1+\epsilon(\rho))\,d\rho$ on $(0,\delta_0]$, so that $\epsilon(\rho)\to0$ as $\rho\to0$. Fix $\delta\in(0,\delta_0]$ and split $m(\lambda)=m_\delta(\lambda)+r_\delta(\lambda)$ into the integrals over $(0,\delta)$ and $[\delta,\rho_{\max}]$. Since $\rho(1+\rho\lambda)^{-2}\le(\rho\lambda^2)^{-1}$, we have $0\le r_\delta(\lambda)\le\nu([\delta,\rho_{\max}])\,\delta^{-1}\lambda^{-2}$. With the substitution $u=\rho\lambda$,
\[
m_\delta(\lambda)=C\lambda^{-(1+\gamma)}\int_0^{\delta\lambda}\frac{u^\gamma}{(1+u)^2}\,\big(1+\epsilon(u/\lambda)\big)\,du .
\]
If $0<\gamma<1$, the integral $\int_0^\infty u^\gamma(1+u)^{-2}du$ is finite and equals $\Gamma(1+\gamma)\Gamma(1-\gamma)=\gamma\pi/\sin(\gamma\pi)$; the contribution of $\epsilon$ is at most $\sup_{\rho<\delta}|\epsilon(\rho)|$ times this integral, and this product tends to zero as $\delta\to0$ because $\epsilon(\rho)\to0$ as $\rho\to0$; and $r_\delta(\lambda)=O(\lambda^{-2})=o(\lambda^{-(1+\gamma)})$. If $\gamma=1$, the integral up to $\delta\lambda$ equals $\log\lambda+O(1)$, and $r_\delta(\lambda)=O(\lambda^{-2})$ is of lower order. If $\gamma>1$, then $\int\rho^{-1}\dnu<\infty$ because the integrand near zero behaves like $C\rho^{\gamma-2}$, and $\lambda^2m(\lambda)\uparrow c^{-1}$ by Lemma~\ref{lem:kernels}.
\end{proof}

\begin{proof}[Proof of Theorem~\ref{thm:exponent}]
By Theorem~\ref{thm:balance}, for $B$ large every direction lies strictly above the floor, so the floor is inactive and the balance condition reads $h_im(\lambda_i^\star)=\mu$ for all $i$, with $\mu\to0$; since $m(\lambda_i^\star)=\mu/h_i\to0$ and $m$ is strictly decreasing, every $\lambda_i^\star\to\infty$.
In case (a), Lemma~\ref{lem:kernels} gives $\lambda^2m(\lambda)\to c^{-1}<\infty$, so $(\lambda_i^\star)^2\sim c^{-1}h_i/\mu$ and $\lambda_i^\star/\lambda_j^\star\to(h_i/h_j)^{1/2}$. In case (b) with $\gamma<1$, Proposition~\ref{prop:asymp} gives $m(\lambda)\sim K\lambda^{-(1+\gamma)}$ with $K>0$, so $(\lambda_i^\star)^{1+\gamma}\sim Kh_i/\mu$ and $\lambda_i^\star/\lambda_j^\star\to(h_i/h_j)^{1/(1+\gamma)}$. For $\gamma=1$, $(\lambda_i^\star)^2/\log\lambda_i^\star\sim Ch_i/\mu$; taking logarithms shows $\log\lambda_i^\star/\log\lambda_j^\star\to1$, and therefore $\lambda_i^\star/\lambda_j^\star\to(h_i/h_j)^{1/2}$ at a logarithmically slow rate. For $\gamma>1$, case (a) applies. In all cases $\lambda_i^\star/\lambda_j^\star\to(h_i/h_j)^{1/\beta}$, and normalizing by $\sum_j\lambda_j^\star=B$ gives \eqref{eq:exponent}.
\end{proof}

\begin{corollary}[Truncated schedules]
\label{cor:truncated}
Let $\nu$ satisfy the hypothesis of Theorem~\ref{thm:exponent}(b) with $\gamma<1$, and let $\nu_{\mathrm{tr}}$ be its restriction to $[\rho_{\min},\rho_{\max}]$, with kernel $m_{\mathrm{tr}}$. Then, for $\lambda\ge\Lambda$,
\[
\Big|\frac{m_{\mathrm{tr}}(\lambda)}{K\lambda^{-(1+\gamma)}}-1\Big|\le \eta(\Lambda)+C'(\lambda\rho_{\min})^{1+\gamma},
\]
where $K=C\gamma\pi/\sin(\gamma\pi)$, $\eta(\Lambda)\to0$ as $\Lambda\to\infty$, and $C'$ depends only on $\nu$. Consequently, the allocation ratios $\lambda_i^\star/\lambda_j^\star$ of the truncated problem are close to $(h_i/h_j)^{1/(1+\gamma)}$ whenever all $\lambda_i^\star$ lie in a range $[\Lambda,\epsilon\rho_{\min}^{-1}]$ with $\Lambda$ large and $\epsilon$ small. For $\lambda\gg\rho_{\min}^{-1}$, $\nu_{\mathrm{tr}}$ satisfies case (a) of Theorem~\ref{thm:exponent} and the square-root law applies.
\end{corollary}
\begin{proof}
We have $0\le m(\lambda)-m_{\mathrm{tr}}(\lambda)\le\int_{(0,\rho_{\min})}\rho\dnu\le C''\rho_{\min}^{1+\gamma}$ for $\rho_{\min}$ small, by the density hypothesis. Dividing by $K\lambda^{-(1+\gamma)}$ and using Proposition~\ref{prop:asymp} for $m$ gives the bound, with $\eta(\Lambda)=\sup_{\lambda\ge\Lambda}|m(\lambda)/(K\lambda^{-(1+\gamma)})-1|$. The statement about ratios follows from the balance condition $h_im_{\mathrm{tr}}(\lambda_i^\star)=\mu$ as in the proof of Theorem~\ref{thm:exponent}. The last statement holds because $\int\rho^{-1}d\nu_{\mathrm{tr}}\le\nu(\R_+)/\rho_{\min}<\infty$.
\end{proof}

\begin{proof}[Proof of Corollary~\ref{cor:cosine}]
With $\rho=\cot^2(\pi t/2)$ we have $t=\frac2\pi\operatorname{arccot}\sqrt\rho$ and $|dt/d\rho|=\pi^{-1}\rho^{-1/2}(1+\rho)^{-1}$. Hence $\nu(d\rho)=w\,|dt/d\rho|\,d\rho=\pi^{-1}\rho^{a-1/2}(1+\rho)^{-1}(1+o(1))\,d\rho$ as $\rho\to0$, which is the hypothesis of Theorem~\ref{thm:exponent}(b) with $C=\pi^{-1}$ and $\gamma=a+\frac12$. The measure is finite because $t\ge t_0$ bounds $\rho$ by $\rho_{\max}=\cot^2(\pi t_0/2)$. By Lemma~\ref{lem:loss}, noise prediction has $w=\rho$ ($a=1$) and data prediction has $w=1$ ($a=0$); velocity prediction has $w=1+\rho=1+o(1)$ as $\rho\to0$ ($a=0$); and min-SNR weighting, which equals $\min(\rho,\gamma_{\mathrm{ms}})$ for some $\gamma_{\mathrm{ms}}>0$ in the data-prediction parameterization \citep{hang2023minsnr}, equals $\rho$ near $\rho=0$ ($a=1$). The statement about truncation is Corollary~\ref{cor:truncated}.
\end{proof}

\begin{proof}[Proof of Proposition~\ref{prop:gain}]
(a) By Lemma~\ref{lem:kernels}, $\lambda\varphi(\lambda)\to c^{-1}$, so $\Rk(\lambda^\iso)=\varphi(B/k)\sum_ih_i\sim c^{-1}k\sum_ih_i/B$. By Theorem~\ref{thm:exponent}(a), $\lambda_i^\star\sim B\sqrt{h_i}/\sum_j\sqrt{h_j}$, so $\Rk(\lambda^\star)=\sum_ih_i\varphi(\lambda_i^\star)\sim c^{-1}\sum_ih_i/\lambda_i^\star\sim c^{-1}(\sum_i\sqrt{h_i})^2/B$.

(b) The lower bound is the Cauchy--Schwarz inequality $(\sum_i\sqrt{h_i})^2\le k\sum_ih_i$, with equality if and only if all $h_i$ are equal. For the upper bound, let $a_i=\sqrt{h_i}\in[a_-,a_+]$ with $a_-=\sqrt{h_k}$ and $a_+=\sqrt{h_1}$. Since $(a_i-a_-)(a_+-a_i)\ge0$, we have $a_i^2\le(a_-+a_+)a_i-a_-a_+$. Summing and writing $x=\frac1k\sum_ia_i\in[a_-,a_+]$ gives $\Gamma(h)\le q(x):=\big((a_-+a_+)x-a_-a_+\big)/x^2$. The function $q$ is maximized at $x=2a_-a_+/(a_-+a_+)$, where $q=(a_-+a_+)^2/(4a_-a_+)$. Since $a_+/a_-=\sqrt\kappa$, this equals $(1+\sqrt\kappa)^2/(4\sqrt\kappa)$.

(c) As $\rho\to0$, $(1+\rho\lambda)^{-1}=1-\rho\lambda+O(\rho^2)$, so $g(\rho)=\rho\big[\sum_ih_i\lambda_i^\star-\frac Bk\sum_ih_i\big]+O(\rho^2)$. By Theorem~\ref{thm:balance}, $\lambda^\star$ and $h$ are similarly ordered, so Chebyshev's sum inequality gives $\frac1k\sum_ih_i\lambda_i^\star\ge\big(\frac1k\sum_ih_i\big)\big(\frac1k\sum_i\lambda_i^\star\big)=\frac{B}{k^2}\sum_ih_i$; the bracket is nonnegative, and if it vanishes, then either all $h_i$ are equal or $\lambda^\star=\lambda^\iso$, in which case $g\equiv0$. As $\rho\to\infty$ with all $\lambda_i^\star>0$ (which holds for $B$ large), $(1+\rho\lambda)^{-1}=(\rho\lambda)^{-1}+O(\rho^{-2})$, so $g(\rho)=\rho^{-1}\big[k\sum_ih_i/B-\sum_ih_i/\lambda_i^\star\big]+O(\rho^{-2})$. By Theorem~\ref{thm:exponent}(a), $B\sum_ih_i/\lambda_i^\star\to(\sum_i\sqrt{h_i})^2\le k\sum_ih_i$, with strict inequality unless all $h_i$ are equal, so the bracket is positive for $B$ large enough when $h$ is not constant, and $g\equiv0$ otherwise.
\end{proof}

\begin{proof}[Proof of Proposition~\ref{prop:local}]
By Lemma~\ref{lem:kernels}, $m$ is positive, strictly decreasing and differentiable, and $m'(\lambda)=-2\int\rho^2(1+\rho\lambda)^{-3}\dnu$, so $\beta_{\mathrm{eff}}(\lambda)=-\lambda m'(\lambda)/m(\lambda)$ is the stated ratio and is positive. It is at most $2$ because $2\lambda\int\rho^2(1+\rho\lambda)^{-3}\dnu=2\int\rho(1+\rho\lambda)^{-2}\frac{\rho\lambda}{1+\rho\lambda}\dnu\le2\int\rho(1+\rho\lambda)^{-2}\dnu$. Differentiating $\log h+\log m(\lambda)=\log\mu$ at fixed $\mu$ gives $d\log h+\frac{d\log m}{d\log\lambda}d\log\lambda=0$, that is $d\log\lambda/d\log h=1/\beta_{\mathrm{eff}}(\lambda)$. Coordinates above the floor satisfy~\eqref{eq:kkt} with the same $\mu$, so integrating this relation from $(h_j,\lambda_j^\star)$ to $(h_i,\lambda_i^\star)$ and bounding $1/\beta_{\mathrm{eff}}$ by $1/\beta_-$ and $1/\beta_+$ gives~\eqref{eq:localexp}. For the limit we argue on the kernels directly, since an asymptotic equivalence for $m$ does not by itself transfer to $m'$. Write $N(\lambda)=-\lambda m'(\lambda)=2\lambda\int\rho^2(1+\rho\lambda)^{-3}\dnu$, so that $\beta_{\mathrm{eff}}=N/m$. If $\int\rho^{-1}\dnu=c^{-1}<\infty$, then $\lambda^2N(\lambda)=2\int\rho^{-1}\big(\rho\lambda/(1+\rho\lambda)\big)^3\dnu\uparrow2c^{-1}$ by monotone convergence, while $\lambda^2m(\lambda)\uparrow c^{-1}$ (Lemma~\ref{lem:kernels}), so $\beta_{\mathrm{eff}}\to2$. Under the hypothesis of Proposition~\ref{prop:asymp} with $\gamma<1$, the substitution $u=\rho\lambda$ used there applies verbatim to $N$ and gives $N(\lambda)\sim2C\lambda^{-(1+\gamma)}\int_0^\infty u^{\gamma+1}(1+u)^{-3}du$, hence
\[
\beta_{\mathrm{eff}}(\lambda)\longrightarrow\frac{2\int_0^\infty u^{\gamma+1}(1+u)^{-3}du}{\int_0^\infty u^{\gamma}(1+u)^{-2}du}=\frac{2\,\Gamma(\gamma+2)\Gamma(1-\gamma)/\Gamma(3)}{\Gamma(\gamma+1)\Gamma(1-\gamma)/\Gamma(2)}=1+\gamma .
\]
For $\gamma=1$ both integrals contribute $\lambda^{-2}\log\lambda$ with the ratio $2$, so $\beta_{\mathrm{eff}}\to2$ at a logarithmically slow rate. In all cases $\beta_{\mathrm{eff}}(\lambda)\to\beta=\min(2,1+\gamma)$.
\end{proof}

\section{Feature distortion of nonlinear decoders}
\label{app:nonlinear}

This appendix relates the risk $\Rk$ to the expected feature distortion $\mathcal D_\rho(M)=\E\|F(S)-F(\Shat)\|^2$ of the nonlinear map $F=\Phi\circ D$.

\begin{assumption}[Regularity]\label{ass:reg}
$J_F$ is $L$-Lipschitz in operator norm on $\R^k$, and $M_2:=(\E\|J_F(S)\|_{\mathrm{op}}^2)^{1/2}<\infty$.
\end{assumption}

\begin{proposition}[Feature distortion]
\label{prop:feature}
Let $S\sim\N(0,I_k)$, let Assumptions~\ref{ass:channel} and~\ref{ass:reg} hold, let $\Shat=\E[S\mid Y]$, and let $\Sigma=\Sigma_\rho(M)$. Then $\mathcal D_\rho(M)=\tr(H\Sigma)+E_\rho$ with
\[
|E_\rho|\le C_1(\tr\Sigma)^{3/2}+C_2(\tr\Sigma)^2,
\]
where $C_1=(2+3^{3/4})\,LM_2\approx4.28\,LM_2$ and $C_2=\big(\tfrac74+3^{3/4}\big)L^2\approx4.03\,L^2$.
\end{proposition}
\begin{proof}
Let $e=S-\Shat$. By Proposition~\ref{prop:lmmse}(b), $e\sim\N(0,\Sigma)$ and $e$ is independent of $\Shat$. Taylor expansion at $\Shat$ gives $F(S)=F(\Shat)+J_F(\Shat)e+r$ with $\|r\|\le\frac L2\|e\|^2$, hence
\[
\|F(S)-F(\Shat)\|^2=e^\top H(\Shat)e+2\big(J_F(\Shat)e\big)^\top r+\|r\|^2 .
\]
We use the Gaussian moment bounds
\begin{equation*}
\begin{aligned}
\E\|e\|^2
&=\tr\Sigma,\\
\E\|e\|^4
&=(\tr\Sigma)^2+2\tr(\Sigma^2)
 \le3(\tr\Sigma)^2,\\
\E\|e\|^3
&\le(\E\|e\|^4)^{3/4}
 \le3^{3/4}(\tr\Sigma)^{3/2}.
\end{aligned}
\end{equation*}
The Lipschitz condition gives
$\|J_F(\Shat)\|_{\mathrm{op}}\le\|J_F(S)\|_{\mathrm{op}}+L\|e\|$,
and hence
\begin{equation*}
\E\|J_F(\Shat)\|_{\mathrm{op}}
\le M_2+L(\tr\Sigma)^{1/2}.
\end{equation*}

\emph{Main term.} By independence, $\E[e^\top H(\Shat)e]=\tr(\E[H(\Shat)]\Sigma)$. Since $\|H(a)-H(b)\|_{\mathrm{op}}\le L\|a-b\|\big(\|J_F(a)\|_{\mathrm{op}}+\|J_F(b)\|_{\mathrm{op}}\big)$, we obtain $\|\E H(\Shat)-H\|_{\mathrm{op}}\le L\,\E\big[\|e\|\big(\|J_F(\Shat)\|_{\mathrm{op}}+\|J_F(S)\|_{\mathrm{op}}\big)\big]\le L(\tr\Sigma)^{1/2}\big(2M_2+L(\tr\Sigma)^{1/2}\big)$, where the term involving $J_F(\Shat)$ factorizes by the independence of $e$ and $\Shat$, while the term involving $J_F(S)$ is bounded by the Cauchy--Schwarz inequality. Because $|\tr(X\Sigma)|\le\|X\|_{\mathrm{op}}\tr\Sigma$ for $\Sigma\succeq0$, the main term deviates from $\tr(H\Sigma)$ by at most $2LM_2(\tr\Sigma)^{3/2}+L^2(\tr\Sigma)^2$.

\emph{Cross term.} By independence, $\E|2(J_F(\Shat)e)^\top r|\le L\,\E\|J_F(\Shat)\|_{\mathrm{op}}\,\E\|e\|^3\le3^{3/4}LM_2(\tr\Sigma)^{3/2}+3^{3/4}L^2(\tr\Sigma)^2$.

\emph{Quadratic term.} $\E\|r\|^2\le\frac{L^2}4\E\|e\|^4\le\frac34L^2(\tr\Sigma)^2$.

Adding the three bounds gives the constants $C_1$ and $C_2$.
\end{proof}

\begin{remark}[Reference point of the distortion]
\label{rem:reference}
Proposition~\ref{prop:feature} measures $\mathcal D_\rho(M)=\E\|F(S)-F(\Shat)\|^2$, the distortion between the clean and the noisy decoded images, whereas the reconstruction metrics of Section~\ref{sec:exp} compare the noisy reconstruction with the original image. Let $X$ denote the original image corresponding to the whitened latent $S$. With $r_\Phi=\Phi(X)-F(S)$ the residual of the clean path and $\Delta_\Phi=F(S)-F(\Shat)$,
\[
\E\|\Phi(X)-F(\Shat)\|^2=\E\|r_\Phi\|^2+\mathcal D_\rho(M)+2\,\E\langle r_\Phi,\Delta_\Phi\rangle .
\]
The first term does not depend on the code, because $G$ is invertible and the clean path is unchanged, but the cross term does. For a linear $F$ with Jacobian $J$ and a residual $r_\Phi=AS$ correlated with the latent, $\E[(S-\Shat)S^\top]=\Sigma_\rho(M)$ gives $2\E\langle r_\Phi,\Delta_\Phi\rangle=2\tr\big(J\Sigma_\rho(M)A^\top\big)$, which varies with the allocation and can take either sign. Parity of clean reconstruction therefore does not make the two evaluations equivalent; reporting the noise-induced distortion relative to the clean reconstruction, integrated against $\nu$, is the measurement that matches the theory.
\end{remark}

\begin{remark}[Non-Gaussian latents]
For non-Gaussian $S$, the LMMSE error is uncorrelated with, but in general not independent of, $\Shat$, so the factorization of the main term fails and the remainder bound above does not apply. The identity between $\Rk$ and the $H$-weighted LMMSE error (Proposition~\ref{prop:lmmse}) remains exact.
\end{remark}

\begin{example}[The surrogate does not determine the nonlinear optimum under uniform weighting]
\label{ex:nonlinear}
Let $S\sim\N(0,I_2)$ and $F(s_1,s_2)=(\tfrac12s_1^2,\,s_2)$, which satisfies Assumption~\ref{ass:reg} on compact sets, and let $M=\diag(\lambda_1,\lambda_2)$. Then $H=\E[J_F(S)^\top J_F(S)]=\diag(\E S_1^2,1)=I_2$, so the surrogate $\Rk$ is minimized by the isotropic allocation for every $\nu$. The exact feature distortion is not: with $v_i=(1+\rho\lambda_i)^{-1}$ and the decomposition $S_1=\Shat_1+e_1$, where $\Shat_1\sim\N(0,1-v_1)$ is independent of $e_1\sim\N(0,v_1)$,
\[
\mathcal D_\rho=\tfrac14\E\big[(2\Shat_1e_1+e_1^2)^2\big]+v_2=\tfrac14\big(4v_1(1-v_1)+3v_1^2\big)+v_2=v_1-\tfrac14v_1^2+v_2 .
\]
Under uniform weighting on the cosine schedule ($\gamma=\tfrac12$), the argument of Proposition~\ref{prop:asymp} applied to $(1+\rho\lambda)^{-1}$ and to $(1+\rho\lambda)^{-2}$ gives $\int v\dnu\sim\lambda^{-1/2}$ and $\int v^2\dnu\sim\tfrac12\lambda^{-1/2}$, so the integrated distortion has leading term $\tfrac78\lambda_1^{-1/2}+\lambda_2^{-1/2}$, whose minimizer under $\lambda_1+\lambda_2=B$ satisfies $\lambda_1^\star/\lambda_2^\star\to(7/8)^{2/3}\approx0.915$ rather than $1$. Under SNR weighting ($\gamma=\tfrac32$) the term $\int v^2\dnu=O(\lambda^{-3/2})$ is of lower order and the ratio tends to $1$, in agreement with the surrogate. On a schedule truncated at $\rho_{\min}$ the ratio returns to $1$ once $\lambda\gg\rho_{\min}^{-1}$, the crossover of Corollary~\ref{cor:truncated}.
\end{example}

\begin{remark}[Scope of the surrogate]
\label{rem:surrogate-scope}
Example~\ref{ex:nonlinear} does not contradict Proposition~\ref{prop:feature}, whose remainder bound holds uniformly in the allocation: the remainder is negligible relative to the leading risk only when the weighting makes $\int(1+\rho\lambda)^{-3/2}\dnu$ of lower order than $\int(1+\rho\lambda)^{-1}\dnu$, which Proposition~\ref{prop:remainder} establishes for SNR weighting and which fails for uniform weighting. The allocation law of Theorem~\ref{thm:exponent} is exact for the LMMSE surrogate under every weighting; its transfer to nonlinear feature distortion is established here only for noise-prediction weighting.
\end{remark}

\begin{proposition}[Integrated remainder]
\label{prop:remainder}
Under the assumptions of Proposition~\ref{prop:feature}, let $M=U\diag(\lambda)U^\top$ with $\lambda_{\min}=\min_i\lambda_i>0$, and let $I_q(\lambda)=\int(1+\rho\lambda)^{-q}\dnu$. Then
\[
\int|E_\rho|\dnu\le C_1k^{3/2}I_{3/2}(\lambda_{\min})+C_2k^2I_2(\lambda_{\min}).
\]
For the cosine schedule with SNR weighting, $\nu(d\rho)=\pi^{-1}\rho^{1/2}(1+\rho)^{-1}d\rho$ on $(0,\rho_{\max}]$, one has $I_{3/2}(\lambda)\le\pi^{-1}\lambda^{-3/2}\big(1+\log(1+\rho_{\max}\lambda)\big)$ and $I_2(\lambda)\le\frac12\lambda^{-3/2}$. Under the allocation of Theorem~\ref{thm:exponent}(a) with fixed $k$, $\lambda_{\min}=\Theta(B)$, so the integrated remainder is $O(B^{-3/2}\log B)$, whereas $\Rk(\lambda^\star)=\Theta(B^{-1})$.
\end{proposition}
\begin{proof}
The first bound follows from $\tr\Sigma_\rho=\sum_i(1+\rho\lambda_i)^{-1}\le k(1+\rho\lambda_{\min})^{-1}$ and Proposition~\ref{prop:feature}. For the cosine measure, the substitution $u=\rho\lambda$ gives $I_{3/2}(\lambda)=\pi^{-1}\lambda^{-3/2}\int_0^{\rho_{\max}\lambda}u^{1/2}(1+u/\lambda)^{-1}(1+u)^{-3/2}du\le\pi^{-1}\lambda^{-3/2}\int_0^{\rho_{\max}\lambda}u^{1/2}(1+u)^{-3/2}du$, and the last integral is at most $\int_0^1du+\int_1^{\max(1,\rho_{\max}\lambda)}u^{-1}du\le1+\log(1+\rho_{\max}\lambda)$. Similarly, $I_2(\lambda)\le\pi^{-1}\lambda^{-3/2}\int_0^\infty u^{1/2}(1+u)^{-2}du=\frac12\lambda^{-3/2}$. Finally, $c^{-1}=\pi^{-1}\int_0^{\rho_{\max}}\rho^{-1/2}(1+\rho)^{-1}d\rho\in(0,1]$, so $\lambda\varphi(\lambda)\to c^{-1}$ (Lemma~\ref{lem:kernels}) and $\lambda_i^\star=\Theta(B)$ give $\Rk(\lambda^\star)=\Theta(B^{-1})$.
\end{proof}

\section{Estimation of the sensitivity matrix}
\label{app:estimation}

\begin{proposition}[Unbiased vector--Jacobian estimator]
\label{prop:vjp}
Let $s_1,\dots,s_N$ be independent copies of $S$, and let $u_{nj}\sim\N(0,I_p)$, $j=1,\dots,n_p$, be independent of each other and of the $s_n$. With $v_{nj}=J_F(s_n)^\top u_{nj}$, the estimator $\hat H=\frac1{Nn_p}\sum_{n,j}v_{nj}v_{nj}^\top$ satisfies $\E\hat H=H$ and requires $Nn_p$ vector--Jacobian products.
\end{proposition}
\begin{proof}
$\E[v_{nj}v_{nj}^\top\mid s_n]=J_F(s_n)^\top\E[u_{nj}u_{nj}^\top]J_F(s_n)=H(s_n)$, and $\E H(s_n)=H$.
\end{proof}

\begin{proof}[Proof of Proposition~\ref{prop:sandwich}]
Since $\varphi(M)\succeq0$, the sandwich implies $a\Rk_H(M)\le\Rk_{\hat H}(M)\le b\Rk_H(M)$ for every feasible $M$. Hence $a\Rk_H(\hat M)\le\Rk_{\hat H}(\hat M)\le\Rk_{\hat H}(M^\star)\le b\Rk_H(M^\star)$, and $\Rk_H>0$ gives the claim. The bound is scale free: replacing $\hat H$ by $c\hat H$ changes neither $\hat M$ nor $b/a$.
\end{proof}

\begin{proposition}[Sample size under a Gaussian model]
\label{prop:sample}
Suppose that $v_1,\dots,v_N$ are independent $\N(0,H)$ vectors and $\hat H=\frac1N\sum_nv_nv_n^\top$. For $t\ge0$ and $\delta=(\sqrt k+t)/\sqrt N<1$, with probability at least $1-2e^{-t^2/2}$, $(1-\delta)^2H\preceq\hat H\preceq(1+\delta)^2H$; hence Proposition~\ref{prop:sandwich} applies with $b/a=\big((1+\delta)/(1-\delta)\big)^2$, which is finite for every $\delta<1$.
\end{proposition}
\begin{proof}
Write $v_n=H^{1/2}g_n$ with $g_n\sim\N(0,I_k)$ and let $\Gamma\in\R^{N\times k}$ have rows $g_n^\top$. Then $\hat H=H^{1/2}(\Gamma^\top\Gamma/N)H^{1/2}$. With the stated probability, all singular values of $\Gamma$ lie in $[\sqrt N-\sqrt k-t,\sqrt N+\sqrt k+t]$ \citep[Corollary~5.35]{vershynin2012intro}, so the eigenvalues of $\Gamma^\top\Gamma/N$ lie in $[(1-\delta)^2,(1+\delta)^2]$, and conjugation by $H^{1/2}$ preserves the PSD order.
\end{proof}

In practice, vector--Jacobian products at different latents are not exactly Gaussian with a common covariance; Proposition~\ref{prop:sandwich} requires only the sandwich condition, which can be checked empirically by comparing estimates from disjoint subsets of probes.

\section{Block-diagonal coding}
\label{app:block}

\begin{proof}[Proof of Lemma~\ref{lem:block}]
\emph{Decomposition.} If $M=\diag(M_1,\dots,M_L)$, then $\varphi(M)=\diag(\varphi(M_1),\dots,\varphi(M_L))$, and $\tr(H\varphi(M))=\sum_b\tr(H_{bb}\varphi(M_b))$ because only the diagonal blocks of $H$ meet the nonzero blocks of $\varphi(M)$. If $M_b=M_0$ for all $b$, then $\sum_b\tr(H_{bb}\varphi(M_0))=\tr(H_{\mathrm{eff}}\varphi(M_0))$.

\emph{Optimal block-diagonal code.} For fixed block spectra, Theorem~\ref{thm:align} applied to each block shows that aligning $M_b$ with the eigenvectors of $H_{bb}$ minimizes $\tr(H_{bb}\varphi(M_b))$, with value $\sum_ih_{b,i}\varphi(\lambda_{b,i})$ where $h_{b,i}$ are the eigenvalues of $H_{bb}$. The budget $\sum_b\tr M_b=\sum_{b,i}\lambda_{b,i}\le B$ couples all blocks linearly, so the problem is \eqref{eq:scalar} over the pooled index set $\{(b,i)\}$, and Theorem~\ref{thm:balance} applies with a single multiplier. For a shared code with budget $L\tr M_0\le B$, the same argument applies to $H_{\mathrm{eff}}$ with budget $B/L$.

\emph{Restriction gap.}
Let $\pin$ be the pinching onto the partition, and let $M^\star$ minimize $\Rk$ over all feasible $M$.
The matrix $\pin(M^\star)$ is block-diagonal and feasible, because $\tr\pin(M^\star)=\tr M^\star$.
Let $H_{\mathrm{OD}}=H-\pin(H)$. By Lemma~\ref{lem:pinch},
\begin{equation*}
\begin{aligned}
\Rk(\pin(M^\star))
&=\tr\big(H\varphi(\pin(M^\star))\big)\\
&\le\tr\big(H\,\pin(\varphi(M^\star))\big)\\
&=\tr\big(\pin(H)\varphi(M^\star)\big)\\
&=\Rk(M^\star)-\tr\big(H_{\mathrm{OD}}\varphi(M^\star)\big).
\end{aligned}
\end{equation*}
The best block-diagonal code has risk at most $\Rk(\pin(M^\star))$.
\end{proof}

This excess-risk bound compares the best code that is block-diagonal, with blocks free to differ, against the unrestricted optimum. Requiring all blocks to share one code $M_0$ is a further restriction whose cost the pinching argument does not control: the shared optimum minimizes $\tr(H_{\mathrm{eff}}\varphi(M_0))$ subject to $\tr M_0\le B/L$, which is in general strictly worse than the pooled per-block optimum. The two coincide exactly when a single $M_0$ realizes the pooled optimum of Lemma~\ref{lem:block} in every block. Identical $H_{bb}$ are sufficient for this but not necessary: for $H_{11}=\diag(100,1)$, $H_{22}=\diag(100,2)$, $\nu=\delta_1$, $\ell=1$ and a total budget of $6$, the pooled solution of Theorem~\ref{thm:balance} is $\lambda^\star=(2,1)$ in both blocks, since the two directions above the floor share the marginal $100/9$ while the floored directions have marginals $1/4$ and $2/4$, both below it; the shared code $M_0=\diag(2,1)$ therefore incurs no additional risk.

The bound depends on the off-block part of $H$, whose magnitude is often summarized by $\|H_{\mathrm{OD}}\|_F^2$. Plug-in estimates of such quadratic functionals are biased upward because $\E\|\hat H_{\mathrm{OD}}\|_F^2=\|H_{\mathrm{OD}}\|_F^2+\E\|\hat H_{\mathrm{OD}}-H_{\mathrm{OD}}\|_F^2$. The following estimator removes this bias.

\begin{proposition}[Unbiased estimation of quadratic functionals of $H$]
\label{prop:ustat}
Let $X_1,\dots,X_n$ be independent, identically distributed, symmetric random matrices with $\E X_j=H$ (for example, per-sample estimates $vv^\top$ from Proposition~\ref{prop:vjp}), and let $A$ be a fixed linear map on symmetric matrices. Then $\frac{1}{n(n-1)}\sum_{j\ne l}\langle A(X_j),A(X_l)\rangle_F$ is an unbiased estimator of $\|A(H)\|_F^2$. In particular, $A(X)=X-\pin(X)$ gives an unbiased estimator of $\|H_{\mathrm{OD}}\|_F^2$.
\end{proposition}
\begin{proof}
For $j\ne l$, independence gives $\E\langle A(X_j),A(X_l)\rangle_F=\langle A(\E X_j),A(\E X_l)\rangle_F=\|A(H)\|_F^2$ by linearity of $A$.
\end{proof}

The bound of Lemma~\ref{lem:block} itself is linear in $H$ for a fixed code, so its plug-in estimate $-\langle\hat H-\pin(\hat H),\varphi(M)\rangle_F$ is unbiased when $M$ is chosen independently of the samples used for $\hat H$.

\section{Implementation}
\label{app:impl}

\paragraph{Deterministic spectral solver.}
On a discrete schedule with $T$ terms, evaluating $m(\lambda)=\sum_t w_t\rho_t(1+\rho_t\lambda)^{-2}$ costs $O(T)$. Algorithm~\ref{alg:rbldc} computes $m^{-1}$ by bisection or interpolation on a precomputed logarithmic grid. The budget equation $\sum_i\lambda_i(\mu)=B$ is solved by bisection in $\log\mu$. A valid finite bracket for $\mu$ is $[\mu_L,\mu_U]$, where $\mu_L=\hat h_1m(B-(k-1)\ell)$ and $\mu_U=\hat h_1m(\ell)$; the corresponding logarithmic bracket is $[\log\mu_L,\log\mu_U]$. Besides one symmetric eigendecomposition of size $k$, nested bisection requires $O(k n_\mu n_\lambda)$ evaluations of $m$, where $n_\mu$ and $n_\lambda$ are the outer and inner iteration counts, giving $O(Tk n_\mu n_\lambda)$ arithmetic cost. Alternatively, precomputing $m$ on $J$ grid points costs $O(JT)$, after which the outer search uses $O(k n_\mu)$ inverse-table lookups. The cost of each lookup and the interpolation error depend on the grid and interpolation method.

\paragraph{Trained parameterization.}
We parameterize the trained code as $G=Q\diag(\sqrt\lambda)P^\top$, with $\lambda=\ell\mathbf 1+(B-k\ell)\,\mathrm{softmax}(u)$. This gives $\tr G^\top G=B$ exactly and $\lambda_i>\ell$. The trained codes use $\ell=0$, so finite softmax parameters give positive eigenvalues in exact arithmetic and hence an invertible code. This parameterization does not impose a fixed positive lower bound on the eigenvalues or a uniform bound on the condition number. An allocation with $\lambda_i=\ell$ can be approached only in the limit. The orthogonal factors are $P=(I-A_P)(I+A_P)^{-1}$ and $Q=(I-A_Q)(I+A_Q)^{-1}$ with skew-symmetric $A_P=(W_P-W_P^\top)/2$ and $A_Q=(W_Q-W_Q^\top)/2$ (the Cayley transform). At $u=0$ and $W_P=W_Q=0$, the parameterization reduces to the isotropic code. By Proposition~\ref{prop:spectrum}, $Q$ does not affect $\Rk$; it is retained for compatibility with joint fine-tuning of the tokenizer. The training objective is a Monte Carlo estimate of $\Rk$ computed with $\hat H$, plus $0.1$ times the variance of $\tilde h_i\,m(\lambda_i)$ over directions with $\lambda_i$ above a small threshold, where $\tilde h_i=p_i^\top\hat Hp_i$ and $p_i$ is the $i$-th column of $P$. By Theorem~\ref{thm:balance}, this penalty vanishes at the optimal allocation. We use Adam \citep{kingma2015adam} with learning rate $5\times10^{-3}$ for 3,000 steps.

\paragraph{Whitening.}
A calibrated whitening transform is fitted on encoder outputs before coding and inverted before decoding. On the primary VA-VAE block-local configuration, it uses the full $k=128$ block covariance. Whitening approximates Assumption~\ref{ass:latent}; the residuals $\|\cov(S)-I_k\|_F/\sqrt k$ are reported in Section~\ref{sec:exp}.

\paragraph{Size and cost.}
The deployed layer is one $k\times k$ matrix per coding block ($k^2$ parameters, for example $16$ for $k=4$ and $16{,}384$ for $k=128$); the trained parameterization additionally stores $u\in\R^k$. Encoding and decoding each add one matrix multiplication per block, and clean reconstruction is unchanged because $G^{-1}G=I_k$.

\section{Numerical evaluation of the theory}
\label{app:numerics}

\paragraph{Allocation exponent.}
Table~\ref{tab:slopes} lists the slopes plotted in Figure~\ref{fig:exponent}. The solver evaluates $m$ on the discretized cosine schedule (1,000 steps, SNRs clipped to $[10^{-5.6},10^{8}]$), solves the allocation problem characterized by Theorem~\ref{thm:balance} by bisection on $\mu$, and reports the least-squares slope of $\log\lambda^\star_i$ against $\log h_i$ over the directions with $\lambda_i^\star>0$ (here $\ell=0$), for $h_i\propto i^{-1}$ and $k=256$. All directions receive positive energy at every budget reported in Table~\ref{tab:slopes}. The larger fitted slopes at small budgets reflect the finite-budget behavior of $m(\lambda)$ before its large-$\lambda$ power-law regime is reached. Proposition~\ref{prop:local} reproduces these values: at $B/k=1$ under SNR weighting the local exponent $1/\beta_{\mathrm{eff}}$ ranges over $[0.545,0.619]$ across the directions that receive energy, which brackets the fitted slope $0.593$, and the realized ratio $\lambda_1^\star/\lambda_k^\star=25.0$ lies inside the bound~\eqref{eq:localexp}, which gives $[20.6,31.0]$.

\begin{table}[htbp]
\caption{Least-squares slope of $\log\lambda_i^\star$ against $\log h_i$ as a function of the budget per direction $B/k$. The last column is the exponent predicted by Corollary~\ref{cor:cosine} for the untruncated schedule.}
\label{tab:slopes}
\centering
\small
\begin{tabular}{@{}lrrrrrrrrc@{}}
\toprule
Weighting & \multicolumn{1}{c}{$10^{-1}$} & \multicolumn{1}{c}{$1$} & \multicolumn{1}{c}{$10$} & \multicolumn{1}{c}{$10^{2}$} & \multicolumn{1}{c}{$10^{3}$} & \multicolumn{1}{c}{$10^{4}$} & \multicolumn{1}{c}{$10^{6}$} & \multicolumn{1}{c}{$10^{8}$} & predicted\\
\midrule
uniform ($z_0$-prediction) & 1.389 & 1.003 & 0.790 & 0.708 & 0.680 & 0.671 & 0.581 & 0.501 & $2/3$\\
SNR ($\varepsilon$-prediction) & 0.641 & 0.593 & 0.543 & 0.516 & 0.505 & 0.502 & 0.500 & 0.500 & $1/2$\\
min-SNR ($\gamma_{\mathrm{ms}}=5$) & 1.047 & 0.681 & 0.561 & 0.520 & 0.507 & 0.502 & 0.500 & 0.500 & $1/2$\\
\bottomrule
\end{tabular}
\end{table}

\paragraph{Learnability of the trained parameterization.}
To test whether the trained parameterization recovers the allocation of Algorithm~\ref{alg:rbldc}, we train a convolutional VAE with $128\times128$ inputs for 50 epochs on 50,000 images from a 100-class subset of ImageNet \citep{deng2009imagenet} (Adam, learning rate $10^{-4}$), freeze its decoder, estimate $\hat H$ from 2,000 latents with four probes each, and train the coding layer as in Appendix~\ref{app:impl}. Table~\ref{tab:learn} reports the relative spectral error $\Delta_\lambda=\|\hat\lambda-\lambda^\star\|/\|\lambda^\star\|$ and the alignment $A_r$ between the leading $r$-dimensional eigenspaces ($r=8$) of the learned and optimal codes (1 for identical subspaces). The relative error lies between $0.0025$ and $0.0043$ in all four settings. These results show close agreement with the reference optimum in the spectrum and the reported leading eigenspaces after 3,000 optimization steps. Algorithm~\ref{alg:rbldc} computes the reference optimum directly, without training the coding layer.

\begin{table}[htbp]
\caption{Recovery of the optimal allocation by the trained parameterization on a small VAE: relative spectral error $\Delta_\lambda$ and leading subspace alignment $A_8$ ($r=8$) after 3,000 steps.}
\label{tab:learn}
\centering
\small
\begin{tabular}{@{}ccrr@{}}
\toprule
$k$ & $\Phi$ & \multicolumn{1}{c}{$\Delta_\lambda$} & \multicolumn{1}{c@{}}{$A_8$}\\
\midrule
64 & identity & 0.00278 & 0.99997\\
64 & random CNN & 0.00247 & 0.99999\\
128 & identity & 0.00426 & 0.99982\\
128 & random CNN & 0.00346 & 0.99997\\
\bottomrule
\end{tabular}
\end{table}

\section{Additional results}
\label{app:results}

\subsection{Reconstruction from noisy latents}

\paragraph{Seed-wise results.}
Table~\ref{tab:seed42} preserves the seed-42 results, including clean-rFID differences and peak pFID improvements. The per-SNR tables and reconstruction figures in this appendix also use seed 42. Across seeds 42--44, the primary configuration has a peak pFID difference of $-108.1\pm2.3$. An additional run with seed 45 on this configuration gives an average pFID difference of $-19.7$ and a peak of $-109.8$; pooling seeds 42--45 gives $-17.7\pm1.8$ and $-108.5\pm2.0$, respectively. The absolute clean-rFID difference is below $0.02$ in all four primary runs. These repetitions use $\Phi$ equal to the identity.

\begin{table}[htbp]
\caption{Seed-42 differences between \rbldc{} and isotropic coding, averaged over 15 SNRs ($\log_{10}\rho\in[-4,3]$), and the largest pFID improvement over the grid with the SNR at which it occurs. Lower is better for pFID and LPIPS; higher is better for PSNR (dB) and SSIM. $\Delta$rFID is computed on clean latents.}
\label{tab:seed42}
\centering
\footnotesize
\setlength{\tabcolsep}{3.2pt}
\begin{tabular}{@{}lrrrrrrr@{}}
\toprule
Tokenizer, coding & \multicolumn{1}{c}{$k$} & \multicolumn{1}{c}{$\Delta$rFID} & \multicolumn{1}{c}{$\Delta$pFID} & \multicolumn{1}{c}{$\Delta$PSNR} & \multicolumn{1}{c}{$\Delta$SSIM} & \multicolumn{1}{c}{$\Delta$LPIPS} & \multicolumn{1}{c@{}}{best $\Delta$pFID ($\log_{10}\rho$)}\\
\midrule
SD-VAE, channel-mix         &   4 & $-0.0002$ & $-1.51$  & $+0.23$ & $+0.0101$ & $-0.0061$ & $-10.53$ ($-2.5$)\\
SD-VAE, block-local         &  16 & $+0.0001$ & $-8.33$  & $+0.63$ & $+0.0268$ & $-0.0208$ & $-38.44$ ($-3.5$)\\
VA-VAE, channel-mix         &  32 & $+0.0007$ & $-1.20$  & $+0.32$ & $+0.0139$ & $-0.0164$ & $-56.85$ ($+1.5$)\\
VA-VAE, block-local         & 128 & $-0.0001$ & $-15.40$ & $+0.81$ & $+0.0312$ & $-0.0372$ & $-105.79$ ($+1.0$)\\
VA-VAE, block-local, $b=4$  & 512 & $-0.0002$ & $-17.04$ & $+1.22$ & $+0.0442$ & $-0.0447$ & $-128.82$ ($+0.5$)\\
\bottomrule
\end{tabular}
\end{table}

\paragraph{Measured sensitivity spectra.}
Table~\ref{tab:measured-H} compares the measured spectra across all five configurations. For the risk ratios in this table, the comparison uses $B=k$ and the kernel
\[
\psi(\lambda)=\Delta x\sum_{j=0}^{255}\frac{\rho_j}{1+\rho_j\lambda},
\qquad x_j=\log\rho_j\ \text{equally spaced in }[-5.6\log 10,8\log 10],
\]
where $\Delta x=(x_{255}-x_0)/255$. This SNR-weighted log-SNR quadrature is used for the spectral comparison and the analytic code variants below; the learned codes for the main reconstruction experiments use the cosine schedule described in Section~\ref{sec:exp}.

Let $\hat h_1\ge\cdots\ge\hat h_k>0$ be the measured eigenvalues and let $\lambda^{\mathrm c}$ minimize $\sum_i\hat h_i\psi(\lambda_i)$ over $\lambda_i\ge0$, $\sum_i\lambda_i=B$. The three reported ratios are
\begin{equation*}
\begin{aligned}
g_{\mathrm{align}}&=\frac{\frac1{10}\sum_{q=1}^{10}\sum_{i,j}\hat h_i(V_q)_{ij}^2\psi(\lambda_j^{\mathrm c})}{\sum_i\hat h_i\psi(\lambda_i^{\mathrm c})},\\
g_{\mathrm{minimax}}&=\frac{\hat h_1\psi(B/k)}{\displaystyle\min_{\lambda_i\ge0,\,\sum_i\lambda_i=B}\max_i\hat h_i\psi(\lambda_i)},
\qquad
g_{\mathrm{pert}}=\frac{k\hat h_1}{\sum_i\hat h_i}.
\end{aligned}
\end{equation*}
The alignment ratio compares ten random orthogonal orientations $V_q$ with the aligned orientation of the same finite-budget optimal spectrum. The rotations are obtained by QR decomposition of Gaussian matrices, using one random stream initialized with seed 0 across configurations. The minimax ratio compares the isotropic and optimal worst-direction integrated risks. For an aligned invertible code, the worst-case amplification of the quadratic metric $\hat H$ under a small coded-latent perturbation is $A(\lambda)=\max_i\hat h_i/\lambda_i$; its optimum is $\sum_i\hat h_i/B$, so $g_{\mathrm{pert}}$ is the isotropic-to-optimal amplification ratio. These quantities differ from the large-budget average-risk gain $\Gamma$ in Proposition~\ref{prop:gain}.

\begin{table}[htbp]
\caption{Measured sensitivity spectra from the seed-42 identity-$\Phi$ estimates. The pFID column repeats the 15-SNR-average differences over seeds 42--44 (mean $\pm$ standard deviation) from Table~\ref{tab:main}. The three ratios are defined above.}
\label{tab:measured-H}
\centering
\footnotesize
\setlength{\tabcolsep}{3pt}
\begin{tabular}{@{}lrrrrrr@{}}
\toprule
Tokenizer, coding & \multicolumn{1}{c}{$k$} & \multicolumn{1}{c}{$\mathrm{cond}(\hat H)$} & \multicolumn{1}{c}{$g_{\mathrm{align}}$} & \multicolumn{1}{c}{$g_{\mathrm{minimax}}$} & \multicolumn{1}{c}{$g_{\mathrm{pert}}$} & \multicolumn{1}{c@{}}{$\Delta$pFID}\\
\midrule
SD-VAE, channel-mix & 4 & 4.0 & $1.21$ & $1.99$ & $2.00$ & $-1.74\pm0.20$\\
SD-VAE, block-local & 16 & 19.8 & $1.50$ & $6.37$ & $6.58$ & $-6.97\pm1.18$\\
VA-VAE, channel-mix & 32 & 114.4 & $2.16$ & $5.22$ & $5.42$ & $-2.59\pm1.28$\\
VA-VAE, block-local & 128 & 408.5 & $3.21$ & $17.16$ & $18.46$ & $-17.07\pm1.60$\\
VA-VAE, block-local, $b=4$ & 512 & 1013.2 & $3.39$ & $48.38$ & $53.38$ & $-22.31\pm4.57$\\
\bottomrule
\end{tabular}
\end{table}

From SD-VAE block-local coding ($k=16$) to VA-VAE channel-mix coding ($k=32$), both the condition number and the alignment ratio increase while the observed pFID improvement decreases. The minimax and perturbation ratios match the ordering of the five observed improvements, including this reversal; this is a descriptive comparison across settings that also change the tokenizer and coding mode.

\paragraph{Per-SNR results.}
Table~\ref{tab:perSNR} lists the values plotted in Figure~\ref{fig:gaps}, and Table~\ref{tab:peaks} lists, for every configuration and metric, the largest improvement over the SNR grid.

\begin{table}[htbp]
\caption{Seed-42 per-SNR differences between \rbldc{} and isotropic coding on the primary configuration (VA-VAE, block-local, $k=128$, $\Phi$ equal to the identity). Entries marked with $^\dagger$ favor isotropic coding (14 of 60).}
\label{tab:perSNR}
\centering
\small
\begin{tabular}{@{}rrrrr@{}}
\toprule
\multicolumn{1}{@{}c}{$\log_{10}\rho$} & \multicolumn{1}{c}{$\Delta$pFID} & \multicolumn{1}{c}{$\Delta$PSNR (dB)} & \multicolumn{1}{c}{$\Delta$SSIM} & \multicolumn{1}{c@{}}{$\Delta$LPIPS}\\
\midrule
$+3.0$ & $-0.121$ & $+0.320$ & $+0.0106$ & $-0.0034$\\
$+2.5$ & $-0.657$ & $+0.819$ & $+0.0268$ & $-0.0156$\\
$+2.0$ & $-3.707$ & $+1.657$ & $+0.0575$ & $-0.0475$\\
$+1.5$ & $-28.740$ & $+2.571$ & $+0.1011$ & $-0.0983$\\
$+1.0$ & $-105.791$ & $+2.946$ & $+0.1230$ & $-0.1308$\\
$+0.5$ & $-70.275$ & $+2.303$ & $+0.0770$ & $-0.0799$\\
$0.0$ & $-35.668$ & $+1.305$ & $+0.0345$ & $-0.0378$\\
$-0.5$ & $+7.670^\dagger$ & $+0.394$ & $+0.0163$ & $-0.0084$\\
$-1.0$ & $+33.166^\dagger$ & $+0.121$ & $+0.0102$ & $+0.0076^\dagger$\\
$-1.5$ & $+2.057^\dagger$ & $+0.286$ & $+0.0092$ & $-0.0428$\\
$-2.0$ & $-3.081$ & $-0.188^\dagger$ & $+0.0032$ & $-0.0811$\\
$-2.5$ & $-9.196$ & $-0.228^\dagger$ & $-0.0005^\dagger$ & $-0.0027$\\
$-3.0$ & $-0.178$ & $-0.101^\dagger$ & $-0.0005^\dagger$ & $-0.0179$\\
$-3.5$ & $-5.582$ & $-0.034^\dagger$ & $-0.0002^\dagger$ & $+0.0016^\dagger$\\
$-4.0$ & $-10.905$ & $-0.012^\dagger$ & $-0.0001^\dagger$ & $-0.0004$\\
\midrule
mean & $-15.40$ & $+0.81$ & $+0.0312$ & $-0.0372$\\
\bottomrule
\end{tabular}
\end{table}

\begin{table}[htbp]
\caption{Largest seed-42 improvement of \rbldc{} over isotropic coding across the 15 SNRs, with the SNR ($\log_{10}\rho$) at which it occurs, for each configuration and metric ($\Phi$ equal to the identity).}
\label{tab:peaks}
\centering
\small
\setlength{\tabcolsep}{4pt}
\begin{tabular}{@{}lrrrrr@{}}
\toprule
Tokenizer and coding & \multicolumn{1}{c}{$k$} & \multicolumn{1}{c}{$\Delta$pFID} & \multicolumn{1}{c}{$\Delta$PSNR (dB)} & \multicolumn{1}{c}{$\Delta$SSIM} & \multicolumn{1}{c@{}}{$\Delta$LPIPS}\\
\midrule
SD-VAE, channel-mix & 4 & $-10.53$ ($-2.5$) & $+0.80$ ($+0.5$) & $+0.032$ ($0.0$) & $-0.037$ ($-1.0$)\\
SD-VAE, block-local & 16 & $-38.44$ ($-3.5$) & $+2.10$ ($+0.5$) & $+0.079$ ($0.0$) & $-0.113$ ($-1.0$)\\
VA-VAE, channel-mix & 32 & $-56.85$ ($+1.5$) & $+1.66$ ($+1.5$) & $+0.070$ ($+1.5$) & $-0.097$ ($-2.0$)\\
VA-VAE, block-local & 128 & $-105.79$ ($+1.0$) & $+2.95$ ($+1.0$) & $+0.123$ ($+1.0$) & $-0.131$ ($+1.0$)\\
VA-VAE, block-local ($b=4$) & 512 & $-128.82$ ($+0.5$) & $+3.47$ ($+1.0$) & $+0.135$ ($+1.0$) & $-0.141$ ($+1.0$)\\
\bottomrule
\end{tabular}
\end{table}

\begin{figure}[htbp]
\centering
\includefig{width=0.95\linewidth}{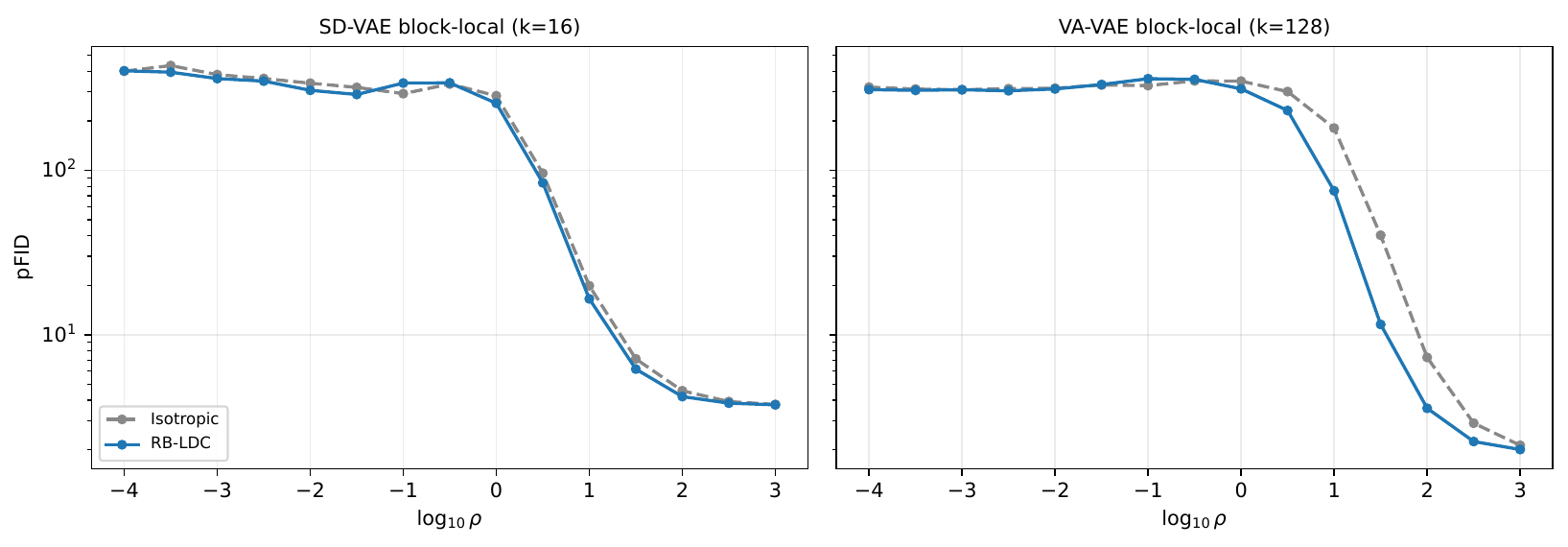}
\caption{Seed-42 absolute pFID of isotropic coding and \rbldc{} ($\Phi$ equal to the identity) for block-local coding on SD-VAE ($k=16$, left) and VA-VAE ($k=128$, right), as a function of $\log_{10}\rho$ on a logarithmic vertical axis.}
\label{fig:backbone}
\end{figure}

\paragraph{Absolute pFID on two tokenizers.}
Figure~\ref{fig:backbone} shows the absolute pFID of both allocations for block-local coding on SD-VAE ($k=16$) and VA-VAE ($k=128$). On SD-VAE, the two curves nearly coincide on the logarithmic scale; on VA-VAE, \rbldc{} is visibly lower between $\log_{10}\rho=+3$ and $0$. The largest difference occurs at $\log_{10}\rho=+1.0$ on VA-VAE ($-105.8$) and at $\log_{10}\rho=-3.5$ on SD-VAE ($-38.4$).

\paragraph{Choice of the feature map.}
On the primary configuration, the eigenvectors of $\hat H$ obtained with the identity, the random convolutional network, and VGG-16 features differ substantially: the best-match overlaps of their eigenvectors are $0.79$ (identity and random CNN), $0.51$ (identity and VGG-16), and $0.54$ (random CNN and VGG-16). Their pFID curves nevertheless lie within approximately 2 points of one another across the SNR grid, so the robustness results are insensitive to the choice among these feature maps in this setting.

\begin{figure}[htbp]
\centering
\includefig{width=0.95\linewidth}{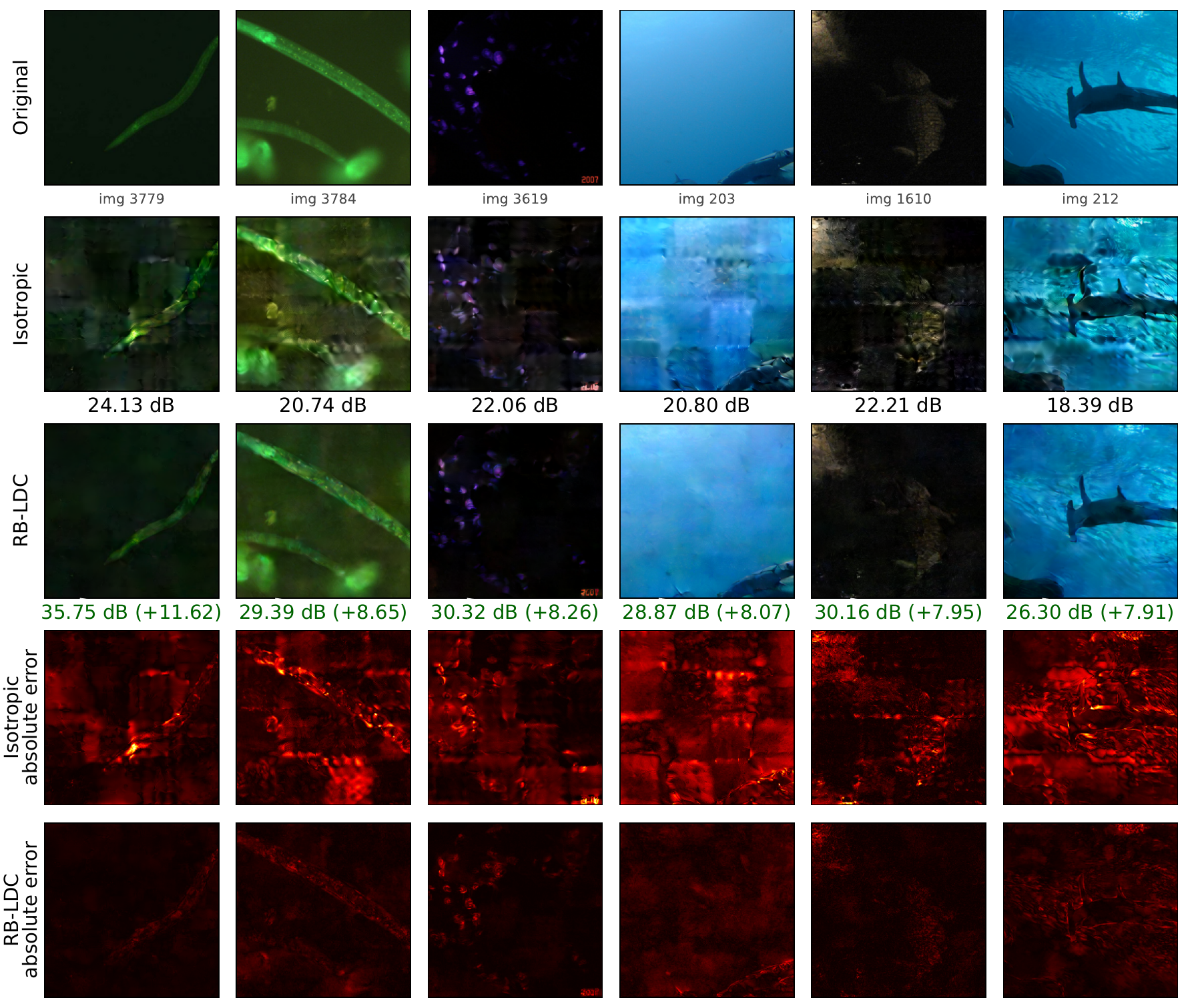}
\caption{Six validation images with the largest per-image PSNR improvement of \rbldc{} over isotropic coding at $\log_{10}\rho=+1$ (VA-VAE, block-local, $b=4$, $k=512$, $\Phi$ equal to the identity). Rows show the originals, the isotropic reconstructions, the \rbldc{} reconstructions, and the absolute errors of both reconstructions. Examples are selected by largest PSNR gain rather than to represent average performance.}
\label{fig:examples}
\end{figure}

\paragraph{Selected examples.}
Figure~\ref{fig:examples} shows six validation images with the largest per-image PSNR improvement at $\log_{10}\rho=+1$ on VA-VAE block-local coding with $b=4$. The images are selected by PSNR improvement, which ranges from $7.91$ to $11.62$~dB among them, and are not representative of the average: the mean PSNR improvement at this SNR is $3.47$~dB (Table~\ref{tab:peaks}).

\paragraph{REPA-E tokenizer.}
We also evaluate the frozen public REPA-E tokenizer \citep{leng2025repae}, \texttt{REPA-E/e2e-vavae-hf}, with a learned identity-$\Phi$ code and block-local coding ($b=2$, $k=B=128$). This reconstruction evaluation uses a fixed subset of 5,000 ImageNet validation images (subset seed 1041), which also supplies the pFID reference. The noisy path uses fp32 arithmetic, run seed 42, and channel-noise base seed 12345, with the same noise shared between the isotropic and coded variants at each SNR. Table~\ref{tab:repae-noisy} gives the full sweep: pFID is lower in all 15 cells, with an unweighted mean difference of $-14.689$ and a largest decrease of $42.576$ at $\log_{10}\rho=0$.

Table~\ref{tab:repae-clean} checks the clean path separately. The fp32 rFID difference is $+0.000442$; with fp64 whitening, coding, inverse recovery, and unwhitening, the difference is zero and all 5,000 saved image pairs are identical. The common decoder remains in fp32 in both evaluations. Comparisons are within each arithmetic setting. 

\begin{table}[htbp]
\caption{REPA-E reconstruction pFID on the fixed 5,000-image validation subset, which also serves as the reference set. The learned identity-$\Phi$ code and isotropic baseline use paired channel noise. Differences are \rbldc{} minus isotropic.}
\label{tab:repae-noisy}
\centering
\small
\begin{tabular}{@{}crrr@{}}
\toprule
$\log_{10}\rho$ & \multicolumn{1}{c}{Isotropic} & \multicolumn{1}{c}{\rbldc{}} & \multicolumn{1}{c@{}}{Difference}\\
\midrule
$-4.0$ & $379.867$ & $375.764$ & $-4.103$\\
$-3.5$ & $371.459$ & $366.420$ & $-5.040$\\
$-3.0$ & $364.674$ & $355.354$ & $-9.319$\\
$-2.5$ & $362.612$ & $326.848$ & $-35.764$\\
$-2.0$ & $343.666$ & $307.511$ & $-36.155$\\
$-1.5$ & $329.293$ & $307.834$ & $-21.458$\\
$-1.0$ & $298.688$ & $270.300$ & $-28.388$\\
$-0.5$ & $266.307$ & $243.181$ & $-23.127$\\
$+0.0$ & $159.585$ & $117.009$ & $-42.576$\\
$+0.5$ & $36.065$ & $24.196$ & $-11.868$\\
$+1.0$ & $7.916$ & $5.990$ & $-1.926$\\
$+1.5$ & $3.191$ & $2.737$ & $-0.453$\\
$+2.0$ & $2.266$ & $2.152$ & $-0.114$\\
$+2.5$ & $2.059$ & $2.023$ & $-0.036$\\
$+3.0$ & $1.997$ & $1.994$ & $-0.003$\\
\midrule
Mean difference & & & $-14.689$\\
\bottomrule
\end{tabular}
\end{table}

\begin{table}[htbp]
\caption{REPA-E clean-rFID check on the same fixed 5,000-image subset. The precision refers to the whitening and coding path; the decoder is fp32 throughout.}
\label{tab:repae-clean}
\centering
\small
\begin{tabular}{@{}lrrr@{}}
\toprule
Arithmetic & \multicolumn{1}{c}{Isotropic} & \multicolumn{1}{c}{\rbldc{}} & \multicolumn{1}{c@{}}{Difference}\\
\midrule
fp32 & $1.976481$ & $1.976924$ & $+0.000442$\\
fp64 coding path & $1.976725$ & $1.976725$ & $0$\\
\bottomrule
\end{tabular}
\end{table}

\subsection{Corruption during sampling and latent transmission}
Both evaluations use a frozen 400k-update LightningDiT-XL/1 checkpoint with VA-VAE at $256\times256$ resolution, CFG 1.0, and Euler sampling on 250 timepoints (249 integration intervals) with timestep shift 0.3. FID compares the resulting sample set with ImageNet validation images using 50,000 reference images and Inception-v3 features. Paired CLIP is the cosine similarity between L2-normalized image embeddings of each recovered image and its corresponding clean generated image, using \texttt{openai/clip-vit-base-patch32}. Images are resized to $224\times224$ with bicubic interpolation and CLIP normalization. At a specified channel SNR $\rho_{\mathrm{ch}}$ (reported in the tables via $\log_{10}\rho_{\mathrm{ch}}$), the added Gaussian channel has parameters
\[
\alpha_{\mathrm{ch}}=\sqrt{\frac{\rho_{\mathrm{ch}}}{1+\rho_{\mathrm{ch}}}},\qquad
\sigma_{\mathrm{ch}}=\frac{1}{\sqrt{1+\rho_{\mathrm{ch}}}}.
\]
All differences below are \rbldc{} minus isotropic, and averages give equal weight to each tested setting. The two experiments use different code variants and sample counts, as specified below.

\paragraph{One-time corruption during sampling.}
This evaluation uses the learned identity-$\Phi$ variant D of Table~\ref{tab:ablation}, with block size $b=2$, $k=B=128$, and isotropic code $I_{128}$. The generator uses the 400k-update EMA checkpoint, with fp32 model and state and TF32 matrix multiplication. For progress $p\in\{0.25,0.50,0.75\}$, the injection index is $\operatorname{round}(p(250-1))$. These percentages count integration intervals, rather than specifying diffusion times $t$. Both codes share the initial noise, class labels, unperturbed sampling prefix, and injected Gaussian noise. We undo the latent mean and standard-deviation normalization at the injection point, whiten the current state, apply the code and channel once, recover with the model-based LMMSE rule of Proposition~\ref{prop:lmmse}, undo whitening, restore the original normalization, and resume the same sampler. This applies the clean-latent recovery rule to an intermediate sampling state. Its LMMSE interpretation at this stage requires the transformed state to have zero mean and identity covariance; Gaussianity is not required.

Each cell contains 5,000 samples, with five per ImageNet class, sample seed 42, and channel-seed base 42,000,000. Table~\ref{tab:injection-full} reports the full paired comparison. FID point estimates are lower for \rbldc{} in all 12 settings, with a mean difference of $-25.530$. Paired CLIP improves in 11 of the 12 settings, with a mean difference of $+0.01921$. At 50\% progress and $\log_{10}\rho_{\mathrm{ch}}=-1$, its difference is $-0.00296$ even though FID improves. For each cell, the CLIP standard error is $s_d/\sqrt{5000}$, where $s_d$ is the sample standard deviation of the per-image paired differences.

\begin{table}[htbp]
\caption{Single-injection results with the learned identity-$\Phi$ code, frozen 400k-update generator, and 5,000 paired samples per cell. The no-injection FID on the matched sample set is $9.612$. CLIP compares each output with its no-injection counterpart; its difference is reported with the paired standard error (SE). Lower FID and higher CLIP are better.}
\label{tab:injection-full}
\centering
\footnotesize
\setlength{\tabcolsep}{2.5pt}
\begin{tabular}{@{}rcrrrrrr@{}}
\toprule
& & \multicolumn{3}{c}{FID} & \multicolumn{3}{c}{Paired CLIP}\\
\cmidrule(lr){3-5}\cmidrule(lr){6-8}
\multicolumn{1}{@{}c}{Progress} & $\log_{10}\rho_{\mathrm{ch}}$ & \multicolumn{1}{c}{Isotropic} & \multicolumn{1}{c}{\rbldc{}} & \multicolumn{1}{c}{$\Delta$} & \multicolumn{1}{c}{Isotropic} & \multicolumn{1}{c}{\rbldc{}} & \multicolumn{1}{c@{}}{$\Delta\pm\mathrm{SE}$}\\
\midrule
25\% & $-1$ & $241.938$ & $191.122$ & $-50.817$ & $0.56400$ & $0.59052$ & $+0.02651\pm0.00068$\\
25\% & $\phantom{+}0$ & $121.488$ & $106.438$ & $-15.050$ & $0.63151$ & $0.64974$ & $+0.01822\pm0.00099$\\
25\% & $+1$ & $14.056$ & $13.395$ & $-0.660$ & $0.84321$ & $0.87109$ & $+0.02788\pm0.00101$\\
25\% & $+2$ & $9.853$ & $9.850$ & $-0.003$ & $0.94012$ & $0.95636$ & $+0.01624\pm0.00066$\\
\midrule
50\% & $-1$ & $323.545$ & $235.477$ & $-88.068$ & $0.61354$ & $0.61058$ & $-0.00296\pm0.00047$\\
50\% & $\phantom{+}0$ & $147.836$ & $124.612$ & $-23.224$ & $0.64823$ & $0.68032$ & $+0.03209\pm0.00095$\\
50\% & $+1$ & $13.438$ & $12.440$ & $-0.998$ & $0.89870$ & $0.91694$ & $+0.01824\pm0.00062$\\
50\% & $+2$ & $9.770$ & $9.710$ & $-0.061$ & $0.97072$ & $0.97878$ & $+0.00805\pm0.00028$\\
\midrule
75\% & $-1$ & $335.773$ & $246.106$ & $-89.667$ & $0.61568$ & $0.62651$ & $+0.01083\pm0.00045$\\
75\% & $\phantom{+}0$ & $179.753$ & $144.848$ & $-34.905$ & $0.66183$ & $0.70760$ & $+0.04576\pm0.00090$\\
75\% & $+1$ & $14.120$ & $11.337$ & $-2.783$ & $0.91968$ & $0.94443$ & $+0.02475\pm0.00047$\\
75\% & $+2$ & $9.804$ & $9.676$ & $-0.127$ & $0.98407$ & $0.98896$ & $+0.00489\pm0.00013$\\
\bottomrule
\end{tabular}
\end{table}

\paragraph{Transmission after generation.}
We first generate 10,000 clean images with the frozen model, ten per ImageNet class, using base seed 42 with two sampling processes (rank seeds 84 and 85). We then re-encode these images with the frozen VA-VAE, whiten their latents, apply a block-local code with $b=2$ and $k=B=128$, pass the result through the Gaussian channel, recover with the LMMSE rule, undo whitening, and decode. The channel grid is $\log_{10}\rho_{\mathrm{ch}}\in\{-4,-3.5,\ldots,+3\}$. For cell index $j=0,\ldots,14$ in this order, the channel seed is $42000+j$, and both codes share the same noise realization.

The analytic allocation for this experiment uses a DINOv2 sensitivity estimate \citep{oquab2023dinov2} and the 256-point SNR-weighted log-SNR kernel $\psi$ defined for Table~\ref{tab:measured-H}. Starting from its KKT spectrum, we clip the eigenvalues to $[c,4c]$, choosing $c$ to preserve the trace budget $B=128$. This cap constrains $\lambda_{\max}/\lambda_{\min}$ to 4, hence $\mathrm{cond}(G)=2$ for the evaluated code. With $L$ the inverse-whitening matrix and $A=\diag(\sqrt\lambda)U^\top$, the weighted-Procrustes orientation is
\[
Q^\star=\arg\min_{Q^\top Q=I}\|QA-L\|_F^2,\qquad G=Q^\star A.
\]
If $AL^\top=V_1\Sigma V_2^\top$, then $Q^\star=V_2V_1^\top$. This fixes the Gram matrix $G^\top G=A^\top A$. Under the isotropic Gaussian channel and LMMSE recovery used in this transmission experiment, changing $Q$ while holding $G^\top G$ fixed leaves the conditional distribution of the recovered latent given the input latent unchanged. 

Table~\ref{tab:transmission-full} reports the results across all 15 SNRs. Sample FID improves at 13 of the 15 SNRs with an average difference of $-7.23$, and paired CLIP improves at 14 SNRs. The FID increases occur at $\log_{10}\rho_{\mathrm{ch}}=-0.5$ and $0$, while CLIP decreases at $-1.5$. The paired reference is the original clean generated image before re-encoding. For this transmission evaluation, FID uses reconstructions quantized to 8 bits, whereas CLIP uses clamped floating-point decoder outputs.

\begin{table}[htbp]
\caption{Post-generation latent transmission of 10,000 samples from the frozen 400k-update generator, using the analytic DINOv2-based code with eigenvalue ratio capped at 4 and weighted-Procrustes orientation ($k=B=128$). The original clean generated sample set has FID $6.4284$. CLIP is paired image similarity to that set. Differences use unrounded values before display rounding.}
\label{tab:transmission-full}
\centering
\footnotesize
\setlength{\tabcolsep}{3.2pt}
\begin{tabular}{@{}crrrrrr@{}}
\toprule
& \multicolumn{3}{c}{FID} & \multicolumn{3}{c}{Paired CLIP}\\
\cmidrule(lr){2-4}\cmidrule(lr){5-7}
$\log_{10}\rho_{\mathrm{ch}}$ & \multicolumn{1}{c}{Isotropic} & \multicolumn{1}{c}{\rbldc{}} & \multicolumn{1}{c}{$\Delta$} & \multicolumn{1}{c}{Isotropic} & \multicolumn{1}{c}{\rbldc{}} & \multicolumn{1}{c@{}}{$\Delta$}\\
\midrule
$-4.0$ & $337.85$ & $331.19$ & $-6.66$ & $0.51603$ & $0.51620$ & $+0.00017$\\
$-3.5$ & $323.74$ & $317.66$ & $-6.08$ & $0.51567$ & $0.51629$ & $+0.00061$\\
$-3.0$ & $314.10$ & $313.62$ & $-0.48$ & $0.51560$ & $0.51870$ & $+0.00310$\\
$-2.5$ & $315.75$ & $312.67$ & $-3.08$ & $0.52085$ & $0.53321$ & $+0.01236$\\
$-2.0$ & $321.18$ & $319.19$ & $-1.98$ & $0.54413$ & $0.55998$ & $+0.01585$\\
$-1.5$ & $340.35$ & $321.76$ & $-18.59$ & $0.55792$ & $0.55671$ & $-0.00121$\\
$-1.0$ & $321.89$ & $317.85$ & $-4.04$ & $0.54582$ & $0.58045$ & $+0.03463$\\
$-0.5$ & $318.12$ & $321.15$ & $+3.03$ & $0.57710$ & $0.57982$ & $+0.00272$\\
$\phantom{+}0.0$ & $322.23$ & $323.59$ & $+1.36$ & $0.57667$ & $0.57898$ & $+0.00231$\\
$+0.5$ & $289.37$ & $272.58$ & $-16.80$ & $0.59193$ & $0.59950$ & $+0.00757$\\
$+1.0$ & $176.30$ & $138.22$ & $-38.09$ & $0.66666$ & $0.69894$ & $+0.03228$\\
$+1.5$ & $46.11$ & $31.91$ & $-14.20$ & $0.82381$ & $0.85644$ & $+0.03263$\\
$+2.0$ & $13.64$ & $11.33$ & $-2.31$ & $0.92760$ & $0.94265$ & $+0.01505$\\
$+2.5$ & $8.46$ & $8.03$ & $-0.42$ & $0.96705$ & $0.97164$ & $+0.00459$\\
$+3.0$ & $7.46$ & $7.36$ & $-0.09$ & $0.97876$ & $0.98014$ & $+0.00139$\\
\bottomrule
\end{tabular}
\end{table}

\subsection{Training on coded representations}
Table~\ref{tab:clean-generation} reports clean-generation FID (gFID) when a LightningDiT-XL/1 model is trained from scratch on each representation, with frozen VA-VAE and a fixed code, except for the denoising-loss variant described in Appendix~\ref{app:ablation}. The training recipe uses seed 42, batch size 256, learning rate $2\times10^{-4}$, and the LightningDiT linear velocity-prediction objective. Sampling uses CFG 1.0 and Euler integration on 250 timepoints (249 intervals), with timestep shift 0.3; gFID uses the 50,000-image ImageNet validation set as the reference. Samples are balanced across 1,000 classes. The 150k and 400k checkpoints use 10K and 50K generated samples, respectively, so comparisons between checkpoints also change the sample count.

The learned generation setting uses CNN-$\Phi$, whereas the sampling-injection experiment uses the primary identity-$\Phi$ code. The dense-$G$ variant corresponds to the unconstrained matrix parameterization trained for $\Rk$ in Table~\ref{tab:ablation}. The two analytic variants use the same DINOv2 sensitivity estimate and capped Gram spectrum as the transmission code above. With $G=Q\diag(\sqrt\lambda)U^\top$, the symmetric choice $Q=U$ gives gFID $12.30$, and the weighted-Procrustes choice gives $11.83$ at 150k updates. Their Gram matrices agree, but their evaluation streams differ: the first uses four sampling workers with seeds 168--171, and the second uses two with seeds 84--85. The main vanilla, isotropic, learned CNN-$\Phi$, and dense-$G$ rows also use four workers with seeds 168--171. Because the weighted-Procrustes run ($11.83$) and the separately evaluated isotropic run ($12.25$) use different worker configurations, their direct numerical ordering spans different sampling streams; the controlled comparison between the two analytic orientations is at 150k updates. All coded rows in the first two panels have higher gFID than vanilla at the corresponding checkpoint.

The final panel uses 50k-update EMA checkpoints and 5K samples, with two sampling workers (seeds 84--85). Its fixed-code baseline trains a new DiT on the primary identity-$\Phi$ representation. The denoising-loss variant updates $G$ jointly with the DiT for the first 3,000 updates and then holds $G$ fixed. Each variant keeps its own latent normalization fixed throughout training: the denoising variant uses statistics of its initial code, and the fixed-code variant uses statistics of the precomputed identity-$\Phi$ code. These runs give gFID $357.88$ and $40.88$, respectively. Appendix~\ref{app:ablation} specifies the denoising objective and its reconstruction results.

\begin{table}[htbp]
\caption{Clean-generation FID with VA-VAE, LightningDiT-XL/1, training seed 42, and CFG 1.0. The panels distinguish the main generation comparison, analytic capped-spectrum variants with different orthogonal factors $Q$, and the separate 50k-update denoising-loss comparison. Sample counts are generated images; the reference contains 50K images throughout.}
\label{tab:clean-generation}
\centering
\small
\setlength{\tabcolsep}{4pt}
\begin{tabular}{@{}lrrr@{}}
\toprule
Representation & \multicolumn{1}{c}{DiT updates} & \multicolumn{1}{c}{Samples} & \multicolumn{1}{c@{}}{gFID}\\
\midrule
Vanilla, no whitening or code & 150k & 10K & $9.07$\\
Whitened isotropic & 150k & 10K & $12.25$\\
Learned CNN-$\Phi$ \rbldc{} & 150k & 10K & $17.93$\\
Dense $G$ for $\Rk$ & 150k & 10K & $21.06$\\
Vanilla, no whitening or code & 400k & 50K & $3.87$\\
Whitened isotropic & 400k & 50K & $6.45$\\
Learned CNN-$\Phi$ \rbldc{} & 400k & 50K & $9.01$\\
Dense $G$ for $\Rk$ & 400k & 50K & $11.42$\\
\midrule
Analytic cap 4, $Q=U$ & 150k & 10K & $12.30$\\
Analytic cap 4, weighted-Procrustes $Q$ & 150k & 10K & $11.83$\\
\midrule
Primary identity-$\Phi$, fixed $G$ & 50k & 5K & $40.88$\\
Denoising loss, $G$ frozen after 3k & 50k & 5K & $357.88$\\
\bottomrule
\end{tabular}
\end{table}

\section{Intermediate-SNR behavior}
\label{app:intermediate}

\paragraph{Magnitudes.}
Table~\ref{tab:intermediate} lists the seed-42 differences at $\log_{10}\rho=-1$, the SNR with the largest pFID increase on the primary configuration, together with the standard deviation of each per-image metric across the 5,000 images. The standard deviations describe the spread of the metrics themselves, not the uncertainty of the paired differences. For reference, the rFID of clean reconstructions on this configuration is $1.91$.

\begin{table}[htbp]
\caption{Seed-42 differences at $\log_{10}\rho=-1$ on the primary configuration. The last column is the standard deviation of the metric across images.}
\label{tab:intermediate}
\centering
\small
\begin{tabular}{@{}lcr@{}}
\toprule
Metric & difference (\rbldc{} $-$ isotropic) & \multicolumn{1}{c@{}}{standard deviation across images}\\
\midrule
pFID & $+33.17\phantom{0}$ & set-level statistic\\
PSNR (dB) & $+\phantom{0}0.12\phantom{0}$ & $4.62$\\
SSIM & $+\phantom{0}0.010$ & $0.143$\\
LPIPS & $+\phantom{0}0.008$ & $0.029$\\
\bottomrule
\end{tabular}
\end{table}

\paragraph{Paired repetitions.}
Table~\ref{tab:valley-paired} summarizes additional seeds 43--45 on the primary identity-$\Phi$ configuration. Their pFID differences are $+30.44$, $+30.24$, and $+28.16$ at $\log_{10}\rho=-1$, the SNR with the largest increase in each run. The standard deviations describe variation across seeds; Table~\ref{tab:intermediate} reports variation across images.

\begin{table}[htbp]
\caption{Intermediate-SNR differences (\rbldc{} minus isotropic) at $\log_{10}\rho=-1$ on the primary configuration with identity-$\Phi$. Values are mean $\pm$ standard deviation across seeds 43--45. Each seed uses the same 5,000 images and noise for both methods. For per-image metrics, the last column is the largest two-sided paired Wilcoxon $p$-value across the three runs. pFID is a set-level statistic, summarized across seeds without a per-image test.}
\label{tab:valley-paired}
\centering
\small
\begin{tabular}{@{}lrr@{}}
\toprule
Metric & \multicolumn{1}{c}{difference across seeds} & \multicolumn{1}{c@{}}{maximum $p$ across seeds}\\
\midrule
pFID & $+29.61\pm1.27$ & --\\
PSNR (dB) & $+0.169\pm0.029$ & $1.22\times10^{-18}$\\
SSIM & $+0.0138\pm0.0004$ & $1.77\times10^{-266}$\\
LPIPS & $+0.00766\pm0.00069$ & $2.93\times10^{-10}$\\
\bottomrule
\end{tabular}
\end{table}

\paragraph{Decomposition of the average.}
For seed 42 on the primary configuration, the pFID differences sum to $+42.89$ over $\log_{10}\rho\in\{-1.5,-1.0,-0.5\}$, to $-204.81$ over $\log_{10}\rho\in\{+0.5,+1.0,+1.5\}$, and to $-69.10$ over the remaining nine SNRs. Dividing their total by 15 gives the average difference of $-15.40$. The configuration with $b=4$ shows the same qualitative pattern, with its largest pFID improvement at $\log_{10}\rho=+0.5$ ($-128.82$).

\paragraph{Direction-dependent shrinkage.}
In the eigenbasis of $M$, the LMMSE signal-retention factor in direction $i$ is $\rho\lambda_i/(1+\rho\lambda_i)$, which changes around $\rho\approx\lambda_i^{-1}$. Under isotropic coding, all directions make this transition at $\rho\approx k/B$. Under \rbldc{}, the transitions occur at different SNRs: directions with larger allocations retain more signal, while those with smaller allocations are shrunk more strongly toward the latent mean. This offers a model-specific interpretation of how the allocation changes reconstructions in this intermediate-SNR regime. The observed PSNR/SSIM improvement and pFID/LPIPS deterioration show that the effects differ across pixel, feature-distance, and distributional metrics. LPIPS compares paired VGG features, so the metric disagreement extends beyond Inception distribution statistics. We relate these observations to the distortion--perception framework of \citet{blau2018perception}; the framework does not predict the SNR location or magnitude of this intermediate-SNR degradation. Its presence and magnitude depend on the allocation spectrum, tokenizer geometry, and evaluation metric.

\section{Component analysis}
\label{app:ablation}

\paragraph{Variants.}
All variants in Table~\ref{tab:ablation} use seed 42, VA-VAE with block-local coding ($k=128$), the same 5,000 images, the same noise realizations, and the same trace budget. Variant A applies identity coding to the tokenizer latent without whitening. Variant B whitens the latent and uses the isotropic code. Variant C whitens the latent and trains a dense matrix $G$ with $\|G\|_F^2=B$ to minimize the Monte Carlo estimate of $\Rk$, with the same optimizer, learning rate, and number of steps as variant D. Variant D is \rbldc{}, which adds the spectral parameterization and the balance penalty of Appendix~\ref{app:impl} to the objective of variant C.

\paragraph{Shares of the improvement.}
The C$\to$D reduction of $7.62$ in 15-SNR-average pFID accounts for $49.5\%$ of the B$\to$D reduction of $15.40$ and $12.6\%$ of the full A$\to$D reduction of $60.27$, including whitening. At $\log_{10}\rho=+1$, C$\to$D reduces pFID by $41.41$; the shares of the full A$\to$D reduction due to whitening, the sensitivity-weighted objective, and the spectral parameterization with the balance penalty are $47\%$, $32\%$, and $21\%$, respectively. Because the analysis of Sections~\ref{sec:setup}--\ref{sec:method} is stated in whitened coordinates, whitening is a prerequisite of the method rather than a contribution.

\paragraph{Balance-penalty ablation.}
Variant D$^-$ uses the same spectral parameterization and softmax trace budget as D, with the balance penalty set to zero. Table~\ref{tab:balance-ablation} compares the codes in a separate matched evaluation. The spectral-recovery indicators, spectrum error $\Delta_\lambda$ and leading subspace alignment $A_8$, remain close to those of D, and the average pFID differences are $-16.251$ and $-16.285$, respectively. Thus, in this setting the spectral parameterization reaches nearly the same allocation and reconstruction performance without the explicit balance penalty.

\begin{table}[htbp]
\caption{Balance-penalty ablation on the primary configuration. D$^-$ and D use training seed 42; all three variants use the same 5,000-image evaluation subset and channel-noise seed 12345. This subset and noise differ from the main evaluation in Table~\ref{tab:ablation}, which uses noise seed 0. pFID differences are relative to the isotropic baseline within this evaluation.}
\label{tab:balance-ablation}
\centering
\small
\begin{tabular}{@{}lrrr@{}}
\toprule
Quantity & \multicolumn{1}{c}{Isotropic} & \multicolumn{1}{c}{D$^-$} & \multicolumn{1}{c@{}}{D}\\
\midrule
Spectrum error $\Delta_\lambda$ & -- & $0.007080$ & $0.005537$\\
Subspace alignment $A_8$ & -- & $0.999857$ & $0.999941$\\
Clean rFID & $1.8206$ & $1.8208$ & $1.8208$\\
15-SNR average $\Delta$pFID & -- & $-16.251$ & $-16.285$\\
Peak $\Delta$pFID ($\log_{10}\rho=+1$) & -- & $-105.848$ & $-106.809$\\
\bottomrule
\end{tabular}
\end{table}

\paragraph{Spectrum reassignment.}
Randomly reassigning the same KKT spectrum across the eigendirections of $H$, under a matched trace budget, increases pFID relative to isotropic coding by $+2.05$ on average over the 15 noisy SNRs and by $+47.3$ at $\log_{10}\rho=+1.0$. The evaluation uses the same SNR grid as the main reconstruction study and includes an additional clean-reconstruction check, which is excluded from this average. This control preserves the spectrum while changing which directions receive the protection energy.

\paragraph{Denoising-loss control.}
We train a code with the same spectral form and 3,000 coding updates as the primary implementation, replacing the sensitivity-weighted risk and balance penalty with the LightningDiT denoising objective. The tokenizer and whitener are frozen, $k=B=128$, and the left orthogonal factor $Q_0$ is fixed to the same initialization. The spectrum starts at $\lambda_i=1$, while the Cayley factors use seed 42 and initialization scale $0.01$; hence the initial Gram matrix is $I_{128}$. The right factor and spectrum are optimized jointly with a new DiT for the first 3,000 updates. The code is then frozen, and the DiT continues to 50k updates for the generation comparison in Table~\ref{tab:clean-generation}.

With $x_1=(G_\phi s-\mu_0)/\sigma_0$, $\varepsilon\sim\N(0,I)$, and $t=\operatorname{sigmoid}(z)$ for $z\sim\N(0,1)$, the linear path and velocity target are $x_t=(1-t)\varepsilon+t x_1$ and $u_t=x_1-\varepsilon$. The objective is
\[
\E\!\left[\operatorname{mean}_{C,H,W}\!\left[(v_\theta(x_t,t,y)-u_t)^2\right]
+\operatorname{mean}_{H,W}\big(1-\cos_C(v_\theta(x_t,t,y),u_t)\big)\right],
\]
where $y$ is the class label and $\cos_C$ is cosine similarity over channels. Gradients reach $G_\phi$ through both $x_t$ and $u_t$. The initial-code statistics $\mu_0,\sigma_0$ remain fixed, including after the code is frozen. The code optimizer is Adam with learning rate $5\times10^{-3}$ and gradient-norm clipping at 10; the DiT uses AdamW with learning rate $2\times10^{-4}$, $\beta=(0.9,0.95)$, zero weight decay, and batch size 256. No sensitivity-risk, balance, or spread-cap term is used.

Table~\ref{tab:denoising-control} evaluates the 3,000-update code by channel corruption and LMMSE recovery on 5,000 validation images (subset seed 1041, channel-noise base seed 0), using the same fixed 5K images as the pFID reference and pairing the noise with an isotropic baseline within this evaluation. The mean $\Delta$pFID is $+78.27$; the largest decrease is $35.23$ at $\log_{10}\rho=-1.5$, while high-SNR pFID increases substantially. The primary identity-$\Phi$ code has mean $\Delta$pFID $-15.40$ in Table~\ref{tab:perSNR}. Because the two evaluations use different 5,000-image validation subsets and reference statistics, their baseline isotropic pFID values differ, and each difference is reported relative to the isotropic baseline of its own evaluation.

\begin{table}[htbp]
\caption{Reconstruction pFID of the denoising-loss code after 3,000 joint coding/DiT updates. Both variants use 5K images and paired noise within this evaluation. Its isotropic values are reported separately from the main evaluation in Table~\ref{tab:perSNR}.}
\label{tab:denoising-control}
\centering
\small
\begin{tabular}{@{}crcc@{}}
\toprule
$\log_{10}\rho$ & \multicolumn{1}{c}{Isotropic} & Denoising-loss $G$ & Difference\\
\midrule
$+3.0$ & $2.228$ & $321.094$ & $+318.866$\\
$+2.5$ & $3.319$ & $315.333$ & $+312.014$\\
$+2.0$ & $8.996$ & $310.966$ & $+301.970$\\
$+1.5$ & $47.984$ & $316.357$ & $+268.373$\\
$+1.0$ & $194.020$ & $339.710$ & $+145.691$\\
$+0.5$ & $316.654$ & $346.315$ & $+\phantom{0}29.660$\\
$+0.0$ & $348.348$ & $326.603$ & $-\phantom{0}21.744$\\
$-0.5$ & $340.998$ & $312.850$ & $-\phantom{0}28.148$\\
$-1.0$ & $326.246$ & $306.720$ & $-\phantom{0}19.526$\\
$-1.5$ & $341.802$ & $306.568$ & $-\phantom{0}35.233$\\
$-2.0$ & $331.886$ & $323.239$ & $-\phantom{00}8.647$\\
$-2.5$ & $329.271$ & $313.676$ & $-\phantom{0}15.595$\\
$-3.0$ & $325.913$ & $295.948$ & $-\phantom{0}29.964$\\
$-3.5$ & $332.091$ & $312.311$ & $-\phantom{0}19.780$\\
$-4.0$ & $343.807$ & $319.931$ & $-\phantom{0}23.876$\\
\midrule
Mean difference & & & $+\phantom{0}78.271$\\
\bottomrule
\end{tabular}
\end{table}

\paragraph{LPIPS.}
LPIPS was not logged during the primary evaluation pass of the ablation variants (variants A and C). Because a subsequent auxiliary pass exhibited minor numerical discrepancies in PSNR and SSIM relative to the primary pipeline, we omit LPIPS from Table~\ref{tab:ablation} and report only metrics computed within the exact same primary evaluation run.

\section{Experimental details}
\label{app:details}

\paragraph{Tokenizers.}
SD-VAE is the Stable Diffusion autoencoder with the fine-tuned decoder released as \texttt{sd-vae-ft-ema} (downsampling factor 8, four latent channels, $32\times32$ latents at $256\times256$). VA-VAE is the released checkpoint \nolinkurl{lavinal712/vavae-imagenet256-f16d32-dinov2} (downsampling factor 16, 32 latent channels, $16\times16$ latents). Both tokenizers are frozen.

\paragraph{Coding configurations.}
Channel-mix coding uses $k=C$ at every spatial position. Block-local coding uses $k=Cb^2$, with $b=2$ for both tokenizers and $b=4$ additionally for VA-VAE. Within each configuration, the same code is applied at every spatial position or block. The budget is $B=k$ per block in all configurations, so the isotropic code is the identity in whitened coordinates.

\paragraph{Sensitivity estimation and training.}
$\hat H$ is estimated with Proposition~\ref{prop:vjp} from 2,000 latents with four probes each. For block-local coding, the VJP outer products are averaged over latents, probes, and block locations to estimate the sensitivity of the shared code (Lemma~\ref{lem:block}). The risk uses SNR weighting on a cosine schedule with 1,000 steps. The coding layer is trained as in Appendix~\ref{app:impl}.

\paragraph{Evaluation.}
We repeat the full pipeline with seeds 42--44 for all five configurations. Table~\ref{tab:main} reports the mean and standard deviation across these runs of each metric difference averaged over the 15 SNRs. We use 5,000 fixed ImageNet validation images at $256\times256$. The SNR grid is $\log_{10}\rho\in\{-4.0,-3.5,\dots,+3.0\}$. At each SNR, the Gaussian noise uses a seed derived from the SNR value and is shared by all allocations. pFID and rFID use Inception-v3 features on the reconstructions against the originals; PSNR, SSIM, and LPIPS (VGG) are averaged over images.

\paragraph{Oracle correction of DiT predictions.}
For Table~\ref{tab:dit-error-correction}, we randomly select 256 ImageNet validation images (seed 42), apply the ADM center crop to $256\times256$, and encode using the posterior mean of frozen VA-VAE. The frozen 400k-update LightningDiT-XL/1 EMA checkpoint uses the images' class labels and CFG 1.0. Its training statistics normalize the raw latent $z$ as $x=(z-\mu_{\mathrm{DiT}})/\sigma_{\mathrm{DiT}}$. At each $t\in\{0.10,0.25,0.50,0.75,0.90\}$, we form $x_t=tx+(1-t)\varepsilon$ with $\varepsilon\sim\N(0,I)$; the predicted velocity $\hat v_t$ gives $\hat x=x_t+(1-t)\hat v_t$. We undo normalization to obtain $\hat z$. Each image and time uses one Gaussian draw, with seeds derived from 42, shared by both corrections.

Corrections use the primary $2\times2$ block whitener, separately from DiT normalization. With $L$ the inverse-whitening matrix, let $e_b=L^{-1}(\hat z_b-z_b)$ for the 64 blocks of dimension $k=128$. Let $P_J$ project onto the top or bottom 16 eigenvectors of the primary identity-$\Phi$ estimate $\hat H$, and set $A_J=\sum_{b=1}^{64}\|P_Je_b\|^2$. For each image and time, the corrected decoder input is defined by
\begin{equation}
\begin{aligned}
B_{\mathrm{corr}}&=\tfrac12\min(A_{\mathrm{high}},A_{\mathrm{low}}),
&\alpha_J&=1-\sqrt{1-B_{\mathrm{corr}}/A_J},\\
\hat z_b^{(J)}&=\hat z_b-L\alpha_JP_Je_b,
&J&\in\{\mathrm{high},\mathrm{low}\}.
\end{aligned}
\label{eq:oracle-correction}
\end{equation}
The reduction in total whitened error energy is $(2\alpha_J-\alpha_J^2)A_J=B_{\mathrm{corr}}$. Rewhitening the corrected latents confirms that the two actual reductions agree to numerical precision.

Image MSE averages squared differences over pixels and channels against the clean decode of $z$, using fp32 decoder outputs without clipping or quantization. Removed-energy percentages average per-image fractions; the reported relative image-MSE reductions are computed from the corresponding mean MSEs.

\paragraph{Compute.}
The main ImageNet reconstruction experiments run on four RTX PRO 6000 Blackwell GPUs with 98~GB of memory each and take between 6 and 9 hours in total, including the learnability study of Appendix~\ref{app:numerics}. The numerical evaluations of the theory (Figure~\ref{fig:exponent}, Table~\ref{tab:slopes}) run on a CPU in a few minutes.

\end{document}